\PassOptionsToPackage{table}{xcolor}
\documentclass[final,5p,times]{elsarticle}

\usepackage{iftex}
\ifPDFTeX\usepackage[utf8]{inputenc}\fi
\usepackage[T1]{fontenc}
\usepackage{url}
\usepackage{xcolor}
\usepackage{graphicx}
\usepackage{amsmath}
\usepackage{amssymb}
\usepackage{enumitem}
\usepackage{multirow}
\usepackage{placeins}
\usepackage{flafter}
\usepackage{array}
\usepackage{longtable}
\usepackage{algorithm}
\usepackage{algpseudocode}
\usepackage{booktabs}
\usepackage{microtype}
\usepackage{xurl}
\usepackage[hidelinks]{hyperref}
\hypersetup{pdfauthor={Yubo Wang, Jingying Ma, Xinliang Zhou, Yangxuan Zhou, Jiquan Wang, Sha Zhao, Yiyuan Yang, Yi Ding, Ziyu Jia, Chenyu Liu, Cuntai Guan}}
\usepackage{etoolbox}
\makeatletter
\pretocmd{\hyper@linkfile}{\let\Hy@href@page\@empty}{}{}
\makeatother
\ifPDFTeX\pdfminorversion=7\else\fi

\providecommand{\KL}{D_{\mathrm{KL}}}
\providecommand{\norm}[1]{\left\lVert #1\right\rVert}
\providecommand{\ms}[2]{$#1${\scriptsize$\pm#2$}}

\journal{Information Fusion}
\biboptions{numbers,sort&compress}
\begin{document}

\begin{frontmatter}

\title{ZeroMAG: Zero-Shot Multimodal Fusion Adapter Generation for Frozen EEG Foundation Models}

\author[inst1,inst2]{Yubo Wang}
\author[inst3]{Jingying Ma}
\author[inst2]{Xinliang Zhou}
\author[inst4]{Yangxuan Zhou}
\author[inst4]{Jiquan Wang}
\author[inst4]{Sha Zhao}
\author[inst5]{Yiyuan Yang}
\author[inst6]{Yi Ding}
\author[inst7]{Ziyu Jia}
\author[inst6]{Chenyu Liu\corref{cor1}}
\ead{chenyu003@ntu.edu.sg}
\author[inst6]{Cuntai Guan\corref{cor1}}
\ead{ctguan@ntu.edu.sg}

\cortext[cor1]{Corresponding authors}
\affiliation[inst1]{op={},organization={Brown University, Providence, RI, USA}}
\affiliation[inst2]{op={},organization={Stanford University, Stanford, CA, USA}}
\affiliation[inst3]{op={},organization={National University of Singapore, Singapore}}
\affiliation[inst4]{op={},organization={Zhejiang University, Hangzhou, China}}
\affiliation[inst5]{op={},organization={Department of Computer Science, University of Oxford, Oxford, United Kingdom}}
\affiliation[inst6]{op={},organization={Nanyang Technological University, Singapore}}
\affiliation[inst7]{op={},organization={Institute of Automation, Chinese Academy of Sciences, Beijing, China}}

\begin{abstract}
Electroencephalography (EEG) recordings often include companion physiological
signals that provide complementary information, yet EEG foundation models
(EFMs) typically expose an EEG-only interface. Extending an existing EFM to
heterogeneous multimodal recordings is challenging when target labels are
unavailable and per-target optimization is impractical. We introduce ZeroMAG,
a zero-shot framework that generates a multimodal fusion adapter for a frozen
EEG encoder and compatible prediction head using unlabeled target recordings
and supplied task context. A configuration-invariant \textit{1+N} adapter combines
the frozen EEG representation with modality-specific temporal--spectral
features through a shared fusion coordinator. It accommodates varying
combinations within a predefined set of supported companion modalities.
A modality--subject--task condition integrates recording-derived modality
and subject information with the supplied task context. Conditional diffusion
then generates adapter parameters in a function-constrained latent space
learned from source adapters, with supervision on both parameters and
induced predictions. Target datasets are excluded from training and model
selection in the ZeroMAG pipeline, and deployment requires neither target
labels nor target-side optimization. Across six held-out target datasets
and three EFM backbones, ZeroMAG improves balanced accuracy by 7.22 percentage
points over EEG-only inference and by 4.89 points over direct weight
regression, while coming within 0.50 points of supervised multimodal
adaptation on average. Ablations support the contributions of learned fusion
and functional supervision. These results support generated adapters as a
means of extending frozen EEG models to heterogeneous multimodal recordings.
\end{abstract}



\end{frontmatter}


\section{Introduction}
\label{sec:intro}

\begin{figure*}[!t]
    \centering
    \includegraphics[width=\textwidth]{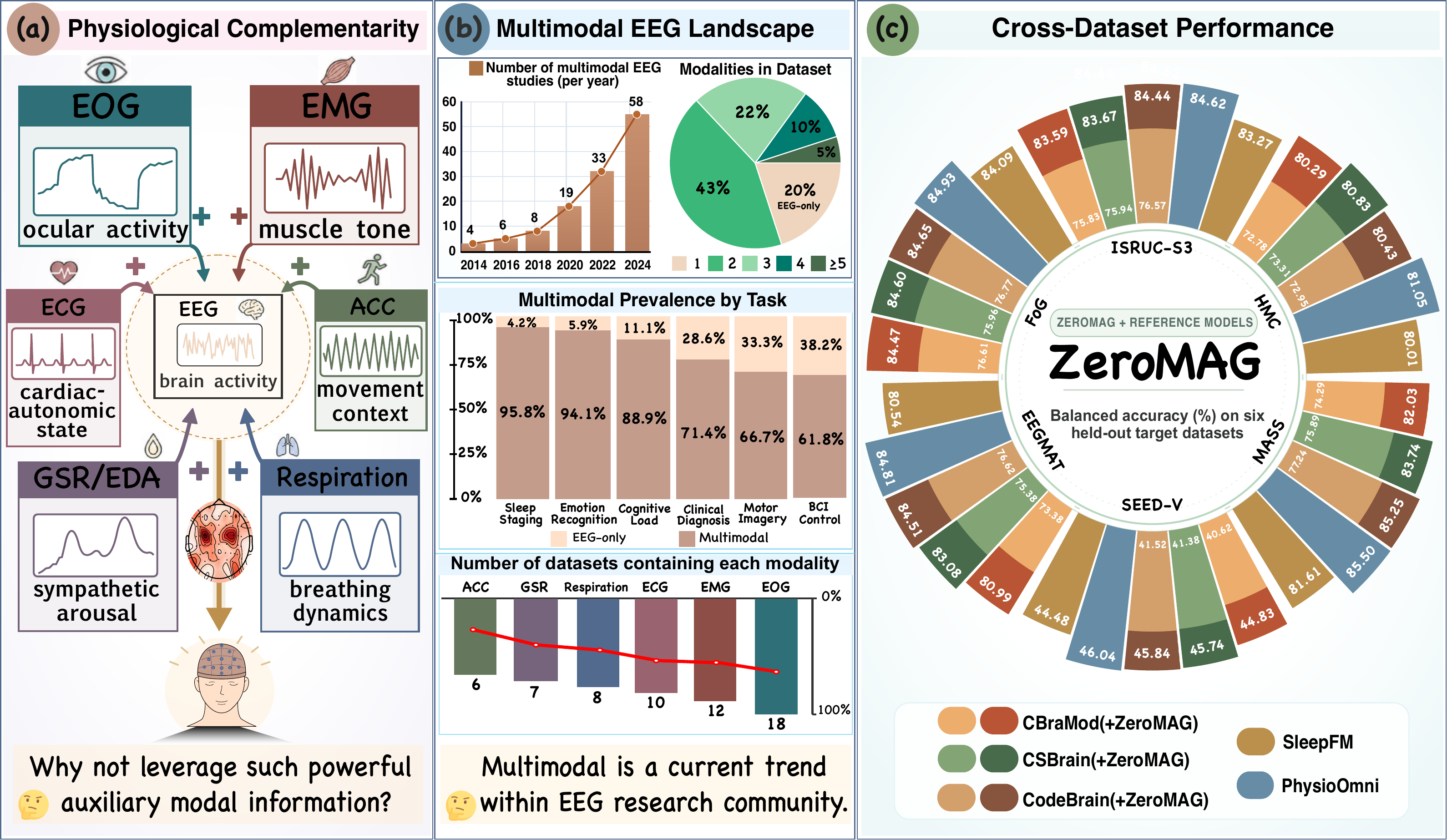}
    \caption{
    Multimodal EEG landscape and representative results.
    (a) Physiological information captured by common companion signals.
    (b) Publication trends, task-wise multimodal prevalence, and companion
    modality availability, based on prior surveys~\citep{lee2025hybrid,yeung2022}.
    (c) Balanced accuracy on six held-out target datasets, comparing three
    EFM backbones with and without ZeroMAG against SleepFM and PhysioOmni.
    }
    \label{fig:companion}
\end{figure*}

A single sensing modality rarely captures the full state of a physiological
system, which is why combining complementary modalities is a central problem
in information fusion~\citep{lahat2015fusion,duan2024biomedical}.
Electroencephalography (EEG) is a clear case. It is a non-invasive and widely
used modality across clinical, cognitive, and brain--computer interface
applications~\citep{lee2025hybrid}, and its large and growing
body of mostly unlabeled recordings has enabled EEG foundation models (EFMs)
that learn reusable representations across downstream
tasks~\citep{cbramod,codebrain,csbrain}. Yet EEG is
increasingly recorded alongside companion signals that capture complementary
physiological information, such as eye movements, muscle activity, and cardiac
dynamics (Fig.~\ref{fig:companion}(a)), and fusing EEG with such peripheral
signals has long been used to improve the recognition of affective and
cognitive states~\citep{zhang2020emotion}.

The demand for such fusion is substantial. A systematic review identified 124 deep learning studies of
hybrid EEG-based multimodal human--computer interfaces published from 2016
to 2024, spanning EEG combined with physiological and neuroimaging
signals~\citep{lee2025hybrid}, and concurrent EEG--fNIRS studies increased
markedly after 2016~\citep{yeung2022}. However, the number and type of
available signals vary widely across datasets
(Fig.~\ref{fig:companion}(b)). A practical fusion system for EEG therefore
cannot assume a fixed modality set: each new dataset arrives with its own
combination of companion signals, which may only partly overlap with the
modalities seen during development.

Existing routes to multimodal EEG fusion each give up something important.
Multimodal physiological and general time-series foundation
models~\citep{sleepfm2024,brantx,physioomni} learn multimodal
representations or signal-alignment components and define their modality
interface during training. Conventional multimodal fusion
networks, including those designed for missing
modalities~\citep{ma2022robust,wang2023shaspec}, learn fusion using labeled training data and are evaluated under
specified missing-modality settings. Adapting an existing EFM with a
fusion module is cheaper, but new recordings often arrive unlabeled because
annotation requires expert review, making supervised fusion for each target
dataset impractical. Label-free adaptation can avoid target labels, but
typically still requires target-side optimization~\citep{shot,tent}. This
raises the central question of this work:

\begin{center}
\begin{minipage}{\linewidth}
\centering\bfseries\itshape
Can multimodal fusion capability be added to an existing EFM
without target labels or target-side training?
\end{minipage}
\end{center}

Answering this question requires addressing three challenges.

\textbf{C1: Heterogeneous modality configurations.}
EEG datasets differ in their companion signals: sleep recordings often include
EOG and EMG, whereas emotion-recognition datasets may include eye movement.
The fusion module must accommodate these differences while allowing every
available signal to contribute complementary information. The challenge is to
fuse varying configurations of supported modalities across datasets within a single
unified framework, without per-dataset redesign.

\textbf{C2: Fusion needs are target-dependent, but labels are often unavailable.}
Even with the same companion signals, the appropriate way to fuse them differs
across datasets, subjects, and prediction tasks, so it must be determined for
each target. However, target labels are usually unavailable because
annotation is costly and requires expert review. The challenge is therefore to
infer target-specific fusion from unlabeled recordings together with the task
context.

\textbf{C3: Parameter reconstruction does not ensure functional consistency.}
A generator of fusion modules is trained to reproduce adapters fitted on
source datasets. Two adapters with similar weights can still make different
predictions, because the output is much more sensitive to some parameters than
to others. A generator trained only to minimize parameter error may therefore
produce adapters that are close to the targets in weight space but fuse the
modalities differently. The challenge is to make the generated adapters agree
with the source adapters in their predictions.

To address these challenges, we propose ZeroMAG, which treats multimodal
fusion as personalized parameter generation: given unlabeled target
recordings and task context, it directly generates a target-specific fusion
adapter while keeping the EFM fixed. Fusion takes place at two levels. At the
feature level, a \textbf{configuration-invariant adapter} fuses reusable
modality-specific branches around a fixed EEG anchor through a shared
coordinator (C1). At the information level, a
\textbf{modality--subject--task condition} integrates modality and subject information derived from unlabeled target
recordings with supplied task context into a single generation condition,
without dataset identifiers (C2).
\textbf{Function-constrained generation} then maps this condition to fusion
parameters through a block-aligned latent space learned from fitted source
adapters, with supervision on the predictions that the generated adapters
induce (C3).

The main contributions of this work are as follows:
\begin{itemize}[leftmargin=*,itemsep=2pt]
\item We formulate zero-shot multimodal fusion for frozen EEG foundation
models: for an unseen target with a varying set of supported companion modalities,
a fusion module is produced from unlabeled recordings and supplied task context, without
target labels or target-side optimization.
\item We design a configuration-invariant \textit{1+N} fusion adapter in which
the frozen EEG representation serves as an invariant anchor and
modality-specific temporal--spectral branches are combined by a shared fusion
coordinator, giving all modality configurations a common parameter structure.
\item We propose a modality--subject--task condition and a function-constrained
latent diffusion generator that together map unlabeled target recordings
and supplied task context to
fusion parameters that agree with reference adapters in their predictions.
\item Across six held-out target datasets and three EFM backbones, ZeroMAG
improves balanced accuracy by 7.22 and 4.89 percentage points over EEG-only
inference and direct weight regression, respectively, comes within 0.50
points of supervised multimodal adaptation on average, and remains
competitive with multimodal foundation models that require multimodal
pretraining (Fig.~\ref{fig:companion}(c)). It also recovers most of the
supervised gain on modality combinations and modality types excluded from
generator training, and learned fusion outperforms mean fusion by 6.58 points
in the nine-dataset ablation average.
\end{itemize}

The remainder of this paper is organized as follows.
Section~\ref{sec:related_work} reviews multimodal information fusion, EEG and
multimodal physiological foundation models, and adaptation and parameter
generation. Section~\ref{sec:preliminaries} formalizes the problem.
Section~\ref{sec:method} presents ZeroMAG. Section~\ref{sec:experiments}
reports the experimental results, and Section~\ref{sec:conclusion} concludes
the paper.

\section{Related Work}
\label{sec:related_work}

\paragraph{Multimodal information fusion}
Combining modalities that observe the same phenomenon from different views is
a long-standing problem in information fusion~\citep{lahat2015fusion}.
Biomedical fusion methods integrate heterogeneous measurements at the data,
feature, or decision level~\citep{duan2024biomedical}. Fusing EEG with
peripheral physiological signals has been studied for emotion and
cognitive-state recognition~\citep{zhang2020emotion} and for hybrid
brain--computer interfaces~\citep{lee2025hybrid}. In physiological emotion
recognition, RHPRNet combines hierarchical fusion with scenario-adapting
pretraining and contrastive alignment~\citep{tang2024rhprnet}.
XSleepFusion uses a dual-stage information bottleneck to reduce intra-modal
redundancy and fuse physiological features for sleep
analysis~\citep{hu2025xsleepfusion}. Complementing these multimodal approaches,
MGIF combines spatial and signal-derived EEG graphs with adaptive gating to
improve BCI reliability~\citep{dai2025mgif}; its fusion operates across EEG
graph representations rather than different physiological signal types.

Modality availability and signal quality can vary across recordings.
A recent survey distinguishes noisy, incomplete, imbalanced, and dynamically
varying multimodal data as separate reliability
challenges~\citep{zhang2026lowquality}. Multimodal transformers are sensitive
to missing modalities, and the best fusion strategy can depend on the
dataset~\citep{ma2022robust}; shared--specific feature modeling addresses
missing modalities within a single model~\citep{wang2023shaspec}.
These missing-modality methods learn fusion using labeled training data and
evaluate robustness to specified missing-modality patterns. ZeroMAG instead
treats the fusion module as the output of a generator: it produces
target-specific fusion parameters for a frozen EEG backbone from unlabeled
recordings and supplied task context, within the supported modality interface.

\paragraph{EEG foundation and multimodal physiological models}
EEG foundation models provide transferable backbones across downstream
tasks~\citep{cbramod,csbrain,codebrain}, but largely retain
EEG-centered interfaces. Prior multimodal physiological work integrates
signals through multimodal foundation-model pretraining or EEG-to-physiology
alignment~\citep{physioomni,sleepfm2024,brantx}: SleepFM and PhysioOmni learn
multimodal physiological representations through pretraining, whereas Brant-X
uses a pretrained EEG foundation model to align EEG with other physiological
signals and trains an EXG encoder. These approaches learn multimodal representations or alignment components
within a specified modality interface. Their support for missing inputs does
not by itself provide a generated adapter for an arbitrary existing EFM. ZeroMAG performs fusion after pretraining, through a generated
plug-in adapter for an existing EFM whose encoder and prediction head remain
frozen.

\paragraph{Adaptation and parameter generation}
Parameter-efficient adaptation methods~\citep{adapters,lora,prompttuning}
still require target-side optimization, and test-time
adaptation~\citep{tent,shot} avoids target labels but still updates
parameters on target data. Hypernetworks generate parameters from task
descriptors or unlabeled
prompts~\citep{hypernetworks,generatedadapters,charakorn2025texttolora,liang2025dnd}.
ZeroMAG instead generates a modular physiological fusion adapter from
unlabeled target information in a structured latent space, and its deployment
requires no pseudo-labels, gradients, or iterative optimization
(Appendix~J.1).

\section{Preliminaries}
\label{sec:preliminaries}

\subsection{EEG Foundation Model}
\label{sec:prelim_efm}

Let $\mathbf{X}^{\mathrm{EEG}}$ denote an EEG sample. An EEG foundation model
consists of an encoder $f(\cdot)$ and a prediction head $h(\cdot)$, so that
\begin{equation}
\mathbf{Z}
=
f(\mathbf{X}^{\mathrm{EEG}}),
\qquad
\mathbf{L}
=
h(\mathbf{Z})
\in\mathbb{R}^{K},
\label{eq:efm}
\end{equation}
where $\mathbf{Z}$ is the EEG representation, $\mathbf{L}$ the corresponding
prediction logits, and $K$ the number of classes.
We directly load the EFM encoder checkpoint and pair it with a compatible
frozen prediction head for the supported task (Section~\ref{sec:frozen_interface}).
Both $f$ and $h$ remain frozen during source adapter fitting, both
generator-training stages, and target inference. 

\subsection{Adapter-Based Multimodal Fusion}
\label{sec:prelim_adapter}

Let $\mathcal{A}$ denote the available companion physiological modalities,
with corresponding signals
$\mathbf{X}^{\mathcal{A}}=\{\mathbf{X}^{m}\}_{m\in\mathcal{A}}$.
To fuse these signals with the EEG representation without modifying the EFM, an adapter
$a(\cdot;\omega)$ is inserted between $f(\cdot)$ and $h(\cdot)$:
\begin{equation}
\mathbf{Z}^{\mathcal{A}}
=
a\left(\mathbf{Z},\mathbf{X}^{\mathcal{A}};\omega\right),
\qquad
\mathbf{L}^{\mathcal{A}}
=
h\left(\mathbf{Z}^{\mathcal{A}}\right),
\label{eq:adapter}
\end{equation}
where $\omega\in\mathbb{R}^{P(\mathcal{A})}$ denotes the adapter parameters,
$\mathbf{Z}^{\mathcal{A}}$ the multimodal representation, and
$\mathbf{L}^{\mathcal{A}}$ the prediction logits returned by the frozen
head $h(\cdot)$.

\subsection{Frozen Prediction Interface}
\label{sec:frozen_interface}
For each EEG backbone, we load the encoder checkpoint and its compatible prediction head, which remains frozen throughout ZeroMAG. EEG-only inference sets the auxiliary residual to zero; multimodal inference adds the adapter-generated residual before the shared head layer. Source labels are used offline to fit reference adapters.

For FoG, we reuse the prediction head supplied by the original method's official implementation and keep it frozen without retraining. FoG is a task-holdout target, so its task factor uses the learned unspecified-task embedding (Appendix~B.6). ZeroMAG adapts supported auxiliary signals, including ACC, to the existing head's input interface (Appendix~H.2).

For the EEG-based interface, given a fused representation $\mathbf{z}^{\mathrm{fused}}=\mathbf{z}^{\mathrm{EEG}}+\Delta\mathbf{z}^{\mathrm{aux}}$, the head is
\begin{align}
\mathbf{u}&=\operatorname{GELU}\!\left(W_s\operatorname{LN}(\mathbf{z}^{\mathrm{fused}})+\mathbf{b}_s\right),\qquad \mathbf{u}\in\mathbb{R}^{128},\\
\boldsymbol{\ell}_t&=W_t\mathbf{u}+\mathbf{b}_t\in\mathbb{R}^{K_t},\qquad
\mathbf{p}_t=\operatorname{softmax}(\boldsymbol{\ell}_t).
\end{align}
The shared projection $(W_s,\mathbf{b}_s)$ is specific to a backbone, as are the task output layers $(W_t,\mathbf{b}_t)$. For a given task $t$, the same output layer is used across its datasets. The adapter and the frozen head therefore share the same prediction path in EEG-only and multimodal evaluation.

We train the generators for CBraMod~\citep{cbramod}, CSBrain~\citep{csbrain}, and CodeBrain~\citep{codebrain} separately. The results across these backbones evaluate separate instantiations of the same framework. They do not represent the transfer of one shared generator checkpoint from one backbone architecture to another. Within each instantiation, the adapter layout, decoder output, frozen EEG representation, and prediction head form the applicable model interface. The block schema is shared within an interface, but the number of blocks and total parameter count depend on the active modalities.

Throughout the appendices, $f,h,e,d,E_{\mathrm{cond}}$, and $s_\theta$ refer to the selected backbone instantiation. The conditioning input remains the recording and task; the selected generator determines the backbone interface.

\subsection{Problem Formulation}
\label{sec:prelim_challenges}

Given a frozen EFM, we aim to adapt it to downstream datasets with a single
framework. These datasets differ in which modalities they record and in the
tasks they target, so each one requires a different adaptation. We formulate
this problem through three objectives.

\noindent\textbf{C1: Define a unified adapter architecture.}
Let $\mathcal{M}$ denote the set of companion physiological modalities. The adapter should maintain a shared internal organization consisting of
modality-specific parameters and shared parameters:
\begin{equation}
\begin{gathered}
\omega
=
\left(
\{\omega^{m}\}_{m\in\mathcal{A}},
\omega^{\mathrm{sh}}
\right),\\
a\left(
\mathbf{Z},
\mathbf{X}^{\mathcal{A}};
\{\omega^{m}\}_{m\in\mathcal{A}},
\omega^{\mathrm{sh}}
\right),
\quad
\forall\,\mathcal{A}\subseteq\mathcal{M}.
\end{gathered}
\label{eq:c1}
\end{equation}
Only the modality-specific components associated with
$m\in\mathcal{A}$ participate for a given configuration, while the overall
adapter organization remains unchanged.

\noindent\textbf{C2: Define a label-free mapping from target data to adapter parameters.}
Let $\mathcal{R}=\{(\mathbf{X}_{i}^{\mathrm{EEG}},\mathbf{X}_{i}^{\mathcal{A}})\}_{i}$
denote the unlabeled target samples, and let $\tau$ denote the prediction
task associated with the frozen EFM.
We seek a generator
\begin{equation}
g:\;(\mathcal{R},\tau)\mapsto\widehat{\omega}\in\mathbb{R}^{P(\mathcal{A})},
\label{eq:c2}
\end{equation}
where $\widehat{\omega}$ denotes the generated adapter parameters.
The generator does not use target labels or dataset identifiers.

\noindent\textbf{C3: Learn the generator from source reference adapters.}
Let $\mathcal{S}$ be a set of source triples $(\mathcal{R},\tau,\omega^{*})$,
where $\omega^{*}$ is a reference adapter fitted on labeled source data.
The generator is trained so that its outputs match these references in both
parameters and predictions through the frozen EFM:
\begin{equation}
\min_{g}\;
\mathbb{E}_{(\mathcal{R},\tau,\omega^{*})\sim\mathcal{S}}
\Big[
\norm{g(\mathcal{R},\tau)-\omega^{*}}^{2}
+
\lambda\,
\delta_{\mathrm{fun}}\!\left(\omega^{*},g(\mathcal{R},\tau)\right)
\Big],
\label{eq:c3}
\end{equation}
where $\delta_{\mathrm{fun}}$ measures the discrepancy between the predictions
induced by two adapters through the same frozen EFM.
At deployment, $g$ produces the target adapter in a single sampling run,
without any gradient updates or target-side optimization.

\begin{figure*}[!t]
    \centering
    \includegraphics[width=\textwidth]{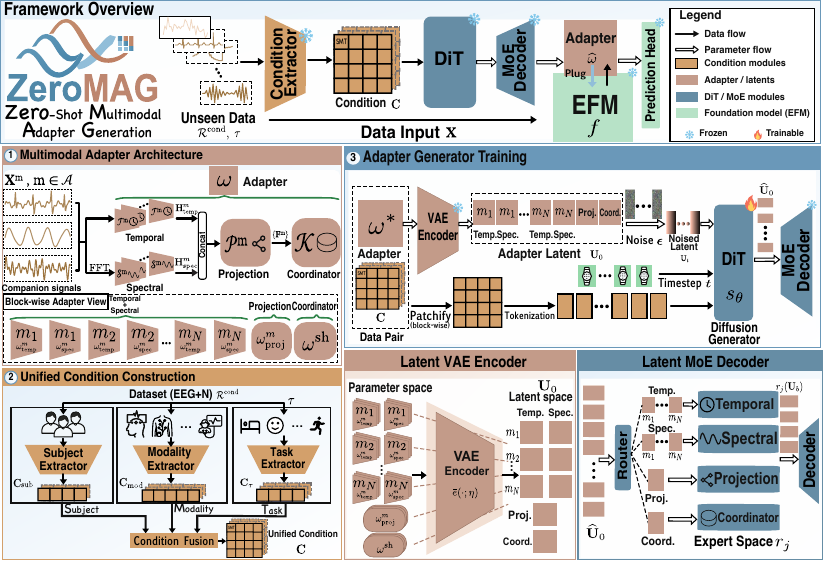}
    \caption{
    Overview of ZeroMAG.
    The framework combines a configuration-invariant multimodal adapter,
    target-derived condition construction, and function-constrained conditional
    generation for zero-shot adapter synthesis.
    }
    \label{fig:zeromag_framework}
\end{figure*}

\section{Methodology}
\label{sec:method}

\subsection{Framework Overview}
\label{sec:framework}

Given an unseen multimodal recording and a supported prediction task,
ZeroMAG generates a plug-in fusion adapter that adds multimodal capability to
a frozen EFM without target labels or target-side optimization.
As illustrated in the upper part of Fig.~\ref{fig:zeromag_framework}, unlabeled target samples
are first encoded into a recording-specific condition.
This condition guides adapter generation in a learned latent space,
and a frozen decoder maps the generated latent representation to adapter
parameters:
\begin{equation}
    \widehat{\omega}
    =
    d\left(
        s_{\theta}\left(
            \boldsymbol{\epsilon},
            E_{\mathrm{cond}}\left(
                \mathcal{R}^{\mathrm{cond}},
                \tau;
                \xi
            \right)
        \right);
        \psi
    \right),
    \qquad
    \boldsymbol{\epsilon}\sim\mathcal{N}(\mathbf{0},\mathbf{I}).
    \label{eq:zeromag_generation}
\end{equation}
Here, $\mathcal{R}^{\mathrm{cond}}$ is the unlabeled conditioning subset,
$\tau$ specifies the task.
The condition extractor $E_{\mathrm{cond}}$, conditional diffusion sampler $s_{\theta}$,
and decoder $d$ are parameterized by $\xi$, $\theta$, and $\psi$,
respectively; $\boldsymbol{\epsilon}$ is the initial Gaussian noise.
The sampler $s_{\theta}$ runs 50 deterministic DDIM steps~\citep{song2020ddim}
with the conditional denoiser $\boldsymbol{\epsilon}_{\theta}$ of
Sec.~\ref{sec:generation}.

The adapter architecture (Sec.~\ref{sec:adapter_arch}) defines what is generated,
the target condition (Sec.~\ref{sec:condition}) specifies what to adapt to, and
the two-stage procedure (Sec.~\ref{sec:generation}) learns a structured adapter
space and conditionally generates within it.

\subsection{Multimodal Adapter Architecture}
\label{sec:adapter_arch}

\textit{Intuition.}
To generate adapters for heterogeneous modality configurations with a single
generator, the adapters must share a common parameter structure. Without it,
each configuration would require its own output space.
This suggests a \textbf{\textit{1+N}} organization, where the frozen EEG representation serves
as the invariant anchor and modality-specific companion branches provide
configuration-dependent extensions.

\textit{Operationalization.}
Following the \textbf{\textit{1+N}} intuition, we treat the frozen EEG representation as the fixed anchor (1) and attach one modality-specific branch for each available companion modality (N).
Regardless of modality type, each companion branch follows the same integration pathway: \textbf{Temporal--Spectral Encoding}, \textbf{Feature Projection}, and \textbf{Shared Coordination}.
The first two components extract and align modality-specific features, while the shared coordinator integrates them with the EEG representation.

\textbf{Temporal--Spectral Encoding.}
For each available companion modality, a temporal encoder extracts patterns
from the signal over time, while a spectral encoder processes its
frequency-domain representation:
\begin{equation}
\begin{aligned}
    \mathbf{H}^{m}_{\mathrm{temp}}
    &=
    \mathcal{T}^{m}\left(
        \mathbf{X}^{m};\omega^{m}_{\mathrm{temp}}
    \right),
    \\
    \mathbf{H}^{m}_{\mathrm{spec}}
    &=
    \mathcal{S}^{m}\left(
        \operatorname{FFT}(\mathbf{X}^{m});\omega^{m}_{\mathrm{spec}}
    \right),
    \qquad m\in\mathcal{A}.
\end{aligned}
\label{eq:temporal_spectral}
\end{equation}
Here, $\mathbf{H}^{m}_{\mathrm{temp}}$ and
$\mathbf{H}^{m}_{\mathrm{spec}}$ are temporal and spectral features.
The encoders $\mathcal{T}^{m}$ and $\mathcal{S}^{m}$ have parameters
$\omega^{m}_{\mathrm{temp}}$ and $\omega^{m}_{\mathrm{spec}}$,
respectively, and $\operatorname{FFT}$ denotes the fast Fourier transform.

\textbf{Feature Projection.}
The temporal and spectral features are concatenated and mapped into the
EFM representation space, providing a common interface for heterogeneous
companion signals:
\begin{equation}
    \mathbf{F}^{m}
    =
    \mathcal{P}^{m}\left(
        \operatorname{Concat}\left(
            \mathbf{H}^{m}_{\mathrm{temp}},
            \mathbf{H}^{m}_{\mathrm{spec}}
        \right);
        \omega^{m}_{\mathrm{proj}}
    \right),
    \qquad m\in\mathcal{A}.
    \label{eq:feature_projection}
\end{equation}
Here, $\mathcal{P}^{m}$ is the projection block with parameters
$\omega^{m}_{\mathrm{proj}}$, and $\operatorname{Concat}$ denotes
feature concatenation.
Its output $\mathbf{F}^{m}$ is the projected feature of modality $m$.

\textbf{Shared Coordination.}
A shared coordinator $\mathcal{K}$ acts as the fusion operator of the
adapter: it jointly processes the frozen EEG representation and the
available companion features and fuses them into the adapted representation:
\begin{equation}
    \mathbf{Z}^{\mathcal{A}}
    =
    \mathcal{K}\left(
        \mathbf{Z},
        \left\{\mathbf{F}^{m}\right\}_{m\in\mathcal{A}};
        \omega^{\mathrm{sh}}
    \right).
    \label{eq:adapter_arch}
\end{equation}
Here, $\omega^{\mathrm{sh}}$ denotes the shared coordinator parameters.
Together, these components define the adapter mapping
$a(\mathbf{Z},\mathbf{X}^{\mathcal{A}};\omega)$. Only available companion branches are evaluated, while the block
schema and functional block identities remain fixed, satisfying C1.

\subsection{Unified Condition Construction}
\label{sec:condition}
\textit{Intuition.}
At deployment, the source-trained generator infers the adapter from unlabeled target recordings and the supported task.
Modality information alone cannot distinguish these cases: recordings with the
same modalities may exhibit different subject-specific physiology, while the same
physiological information may serve different prediction tasks.
This suggests conditioning adapter generation on modality, subject, and task
information.

\textit{Operationalization.}
Following this intuition, the condition encoder $E_{\mathrm{cond}}$ builds one
factor for each source of variation (modality, subject, and task) and combines
them through a fusion module $F_{\mathrm{cond}}$:
\begin{equation}
\mathbf{C}
=
E_{\mathrm{cond}}\left(
\mathcal{R}^{\mathrm{cond}},\tau;\xi
\right)
=
F_{\mathrm{cond}}\left(
\left[
\mathbf{C}_{\mathrm{mod}};
\mathbf{C}_{\mathrm{sub}};
\mathbf{C}_{\tau}
\right]
\right).
\label{eq:condition}
\end{equation}
Here, $\mathcal{R}^{\mathrm{cond}}$ denotes the unlabeled target subset used
for condition construction, $\tau$ specifies the prediction task, and
$\mathbf{C}$ is the resulting target condition.
$\mathbf{C}_{\mathrm{mod}}$, $\mathbf{C}_{\mathrm{sub}}$, and
$\mathbf{C}_{\tau}$ denote the modality, subject, and task factors,
respectively, while $\xi$ contains all trainable parameters of
$E_{\mathrm{cond}}$.

\textbf{Factor Construction.}
Let $i$ index samples in $\mathcal{R}^{\mathrm{cond}}$.
For each modality $m\in\{\mathrm{EEG}\}\cup\mathcal{A}$,
$\mathbf{V}_{i}^{m}$ denotes its sample-level feature and
$\mathbf{V}^{m}$ the corresponding recording-level feature.
The three factors are constructed as
\begin{equation}
\begin{aligned}
\mathbf{C}_{\mathrm{mod}}
&=
F_{\mathrm{mod}}\!\left(
\left\{
\mathbf{V}^{m}+\mathbf{e}_{\mathrm{mod}}^{m}
\right\}_{m\in\{\mathrm{EEG}\}\cup\mathcal{A}}
\right),\\
\mathbf{C}_{\mathrm{sub}}
&=
F_{\mathrm{sub}}\!\left(
\mathbf{q}_{\mathrm{sub}},
\left\{
\mathbf{V}_{i}^{m}
\right\}_{m\in\{\mathrm{EEG}\}\cup\mathcal{A},\,i}
\right),\\
\mathbf{C}_{\tau}
&=
E_{\mathrm{task}}(\tau).
\end{aligned}
\label{eq:condition_factors}
\end{equation}
Here, $F_{\mathrm{mod}}$ is permutation-invariant set pooling~\citep{deepsets}
over recording-level features tagged with learned modality embeddings
$\mathbf{e}_{\mathrm{mod}}^{m}$; $F_{\mathrm{sub}}$ is attention
pooling of all sample-level features with a learned,
task-independent query $\mathbf{q}_{\mathrm{sub}}$; and $E_{\mathrm{task}}$ is
a learned embedding of the supported task. All are part of $\xi$; details and
invariance properties are given in Appendix~B.

\subsection{Adapter Generator Training}
\label{sec:generation}

\textit{Intuition.}
Fitted reference adapters reveal how heterogeneous blocks can be jointly
configured to work with the frozen EFM.
Their shared architecture and fitting objective suggest that these
configurations contain recurring structure, making it unnecessary to model
every degree of freedom in the raw weight space.
Learning a compact, block-aligned representation can capture this structure
while retaining the distinct roles and dependencies of adapter components.
Conditional generation can then focus on variations within this learned
representation, while functional supervision keeps both the decoded and the
generated adapters in predictive agreement with their references.

\textit{Operationalization.}
We implement this design in two stages: function-preserving representation
learning followed by function-constrained conditional generation.
Both use the functional discrepancy $\delta_{\mathrm{fun}}$ introduced
in C3, which we evaluate on a source subset $\mathcal{R}^{\mathrm{fun}}$
held out from reference-adapter fitting:
\begin{equation}
\delta_{\mathrm{fun}}(\omega,\omega')
=
\frac{1}{|\mathcal{R}^{\mathrm{fun}}|}
\sum_{\mathbf{r}_{i}\in\mathcal{R}^{\mathrm{fun}}}
D_{\mathrm{KL}}\!\left(
\sigma(\mathbf{L}_{i}^{\mathcal{A}}(\omega))
\,\middle\|\,
\sigma(\mathbf{L}_{i}^{\mathcal{A}}(\omega'))
\right),
\label{eq:functional_discrepancy}
\end{equation}
where $\mathbf{L}_{i}^{\mathcal{A}}(\omega)$ denotes the logits of sample
$\mathbf{r}_{i}$ produced with adapter parameters $\omega$, and $\sigma$
denotes softmax.
This label-free measure compares two adapters solely through the predictions
they induce in the same frozen EFM.


\subsubsection{Stage 1: Function-Preserving Representation Learning}

\textbf{Block-Wise Latent Encoding.}
Let $\omega^{*}=[\omega^{*}_{b}]_{b=1}^{B}$ denote a fitted reference adapter
partitioned into the functional blocks of Sec.~\ref{sec:adapter_arch}.
A VAE encoder maps each block to a latent token
$\mathbf{U}_{b,\mathrm{pre}}$, which is combined with a learned block
embedding $\mathbf{E}_{b}$ and contextualized by a cross-block transformer
(CBT):
\begin{equation}
\mathbf{U}
=
\operatorname{CBT}\!\left(
[\mathbf{U}_{b,\mathrm{pre}}+\mathbf{E}_{b}]_{b=1}^{B}
\right),
\label{eq:cross_block}
\end{equation}
whose $b$-th token is denoted $\mathbf{U}_{b}$.
We denote the resulting stochastic encoder by $e(\cdot;\eta)$.

\textbf{MoE Decoding.}
To share capacity across heterogeneous block types while respecting their
distinct output forms, each token is routed to shared experts
$\{r_j\}$ according to its content and block identity, followed by a
block-type-specific output head:
\begin{equation}
\begin{aligned}
\boldsymbol{\pi}_{b}
&=
\operatorname{softmax}\!\left(
\operatorname{route}(
[\mathbf{U}_{b};\mathbf{E}_{b}];\psi_r)
\right),\\
\widetilde{\omega}_{b}
&=
\operatorname{head}_{\operatorname{type}(b)}
\!\left(
\sum_{j\in\operatorname{TopK}(\boldsymbol{\pi}_{b})}
\pi_{b,j}r_j(\mathbf{U}_{b})
\right),
\end{aligned}
\label{eq:moe_decoder}
\end{equation}
where
$\operatorname{type}(b)\in
\{\mathrm{temp},\mathrm{spec},\mathrm{proj},\mathrm{shared}\}$.
Collectively, the decoder gives $\widetilde{\omega}=d(\mathbf{U};\psi)$.

\textbf{Function-Preserving Reconstruction.}
Since parameter reconstruction alone does not guarantee that adapter behavior
is preserved, Stage~1 combines parameter-level and function-level objectives:
\begin{equation}
\mathcal{L}_{\mathrm{AE}}
=
\mathcal{L}_{\mathrm{blk}}
+
\lambda_{\mathrm{cos}}\mathcal{L}_{\mathrm{cos}}
+
\lambda_{\mathrm{fun}}
\delta_{\mathrm{fun}}(
\omega^{*},\widetilde{\omega})
+
\beta\mathcal{L}_{\mathrm{KL}}
+
\lambda_{\mathrm{bal}}\mathcal{L}_{\mathrm{bal}},
\label{eq:ae_loss}
\end{equation}
where $\mathcal{L}_{\mathrm{blk}}$ and $\mathcal{L}_{\mathrm{cos}}$
reconstruct adapter parameters, $\delta_{\mathrm{fun}}$ preserves predictive
behavior, $\mathcal{L}_{\mathrm{KL}}$ regularizes the latent posterior, and
$\mathcal{L}_{\mathrm{bal}}$ balances expert utilization.
After training, the encoder and decoder are frozen, and
$\mathbf{U}_{0}=\overline{e}(\omega^{*};\eta)$, obtained by contextualizing
posterior means, serves as the deterministic target latent for Stage~2.

\subsubsection{Stage 2: Function-Constrained Conditional Generation}

\textbf{Conditional Latent Diffusion.}
For each source triple $(\mathcal{R},\tau,\omega^{*})$, a condition
$\mathbf{C}=E_{\mathrm{cond}}(\mathcal{R}^{\mathrm{cond}},\tau;\xi)$ is
constructed from an unlabeled subset
$\mathcal{R}^{\mathrm{cond}}\subseteq\mathcal{R}^{\mathrm{fit}}$, where
$\mathcal{R}^{\mathrm{fit}}$ is the subset used to fit $\omega^{*}$.
In the frozen Stage~1 latent space, a DiT denoiser~\citep{peebles2023scalable}
$\boldsymbol{\epsilon}_{\theta}(\mathbf{U}_{t},t,\mathbf{C})$ predicts the
noise added to $\mathbf{U}_{0}$.
Its prediction yields the clean-latent estimate
$\widehat{\mathbf{U}}_{0}
=(\mathbf{U}_{t}-\sqrt{1-\bar{\alpha}_{t}}\,
\boldsymbol{\epsilon}_{\theta})/\sqrt{\bar{\alpha}_{t}}$,
which the frozen decoder maps to
$\widehat{\omega}=d(\widehat{\mathbf{U}}_{0};\psi)$.

\textbf{Function-Constrained Generation.}
The condition encoder and denoiser are jointly optimized by
\begin{equation}
\mathcal{L}_{\mathrm{gen}}
=
\mathcal{L}_{\mathrm{diff}}
+
\lambda_{\mathrm{lat}}\mathcal{L}_{\mathrm{lat}}
+
\lambda_{\mathrm{w}}\mathcal{L}_{\mathrm{w}}
+
\lambda_{\mathrm{gf}}
\delta_{\mathrm{fun}}(
\omega^{*},\widehat{\omega}),
\label{eq:generation_loss}
\end{equation}
where $\mathcal{L}_{\mathrm{diff}}$ supervises denoising,
$\mathcal{L}_{\mathrm{lat}}$ and $\mathcal{L}_{\mathrm{w}}$ align
$\widehat{\mathbf{U}}_{0}$ and $\widehat{\omega}$ with their references, and
$\delta_{\mathrm{fun}}$ aligns predictive behavior, with gradients passing
through the frozen decoder.
Because the division by $\sqrt{\bar{\alpha}_{t}}$ amplifies noise-prediction
errors at high noise levels, the latent, weight, and functional terms are
applied only at low-noise timesteps (Appendix~H.1), whereas
$\mathcal{L}_{\mathrm{diff}}$ is applied at all timesteps.

After training, $E_{\mathrm{cond}}$, $\boldsymbol{\epsilon}_{\theta}$, and $d$
are fixed and realize the label-free mapping of C2 (Eq.~\eqref{eq:zeromag_generation}).

\subsection{Training and Deployment Protocol}
\label{sec:training_protocol}
\begin{figure*}[t]\centering
\includegraphics[width=.82\textwidth]{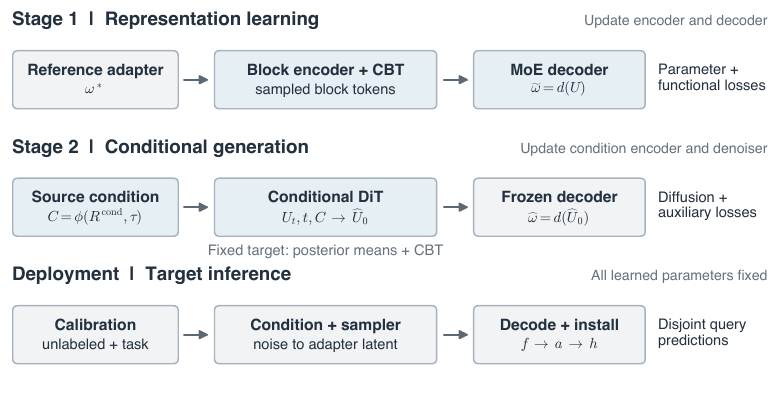}
\caption{Training and deployment within one separately trained backbone instantiation. Stage 1 produces the encoder and decoder. Stage 2 uses a fixed canonical latent and propagates auxiliary-loss gradients through the frozen decoder. Deployment uses unlabeled calibration data and fixed model parameters.}\label{fig:app_workflow}
\end{figure*}

ZeroMAG has two learned stages followed by deployment. Figure~\ref{fig:app_workflow}
makes the update boundaries explicit. Stage~1 learns the adapter representation
and decoder. Stage~2 trains the condition encoder and diffusion denoiser while
retaining the Stage~1 encoder and decoder. A frozen decoder still propagates
auxiliary-loss gradients to the generated latent; frozen parameters do not
imply a zero input Jacobian. Deployment is an inference procedure: only the
unlabeled calibration subset and supplied task context are used, and all
model parameters remain fixed. Detailed algorithms and objective analysis
are provided in Appendices~D
and~E.
\begin{table}[!t]\centering\footnotesize
\setlength{\tabcolsep}{3pt}
\caption{Task coverage and holdout type of the target datasets.}\label{tab:dataset_tasks}
\begin{tabular}{lll}\toprule Target & Task & Holdout\\\midrule
ISRUC-S3 & Sleep staging & Dataset\\HMC & Sleep staging & Dataset\\MASS & Sleep staging & Dataset\\
SEED-V & Emotion recognition & Dataset\\EEGMAT & Cognitive-load estimation & Dataset\\FoG & Freezing-of-gait detection & Task\\\midrule
CAP & Sleep staging & Dataset\\DREAMER & Emotion recognition & Dataset\\BCI-IV-2a (BCI-IV) & Motor imagery & Dataset\\\bottomrule\end{tabular}\end{table}
\begin{table*}[t]
\centering\small
\setlength{\tabcolsep}{4pt}
\renewcommand{\arraystretch}{1.05}
\caption{Implementation settings. $M$ is the number of active companion modalities.
Auxiliary Stage~2 losses share the same low-noise gate; diffusion loss is always active.}
\label{tab:implementation}
\begin{tabular}{@{}lll@{}}
\toprule
Group & Setting & Value \\
\midrule
Representation & Active parameter blocks / latent shape & $3M+1$ / $(3M+1)\times128$ \\
 & CBT layers / hidden width / heads & $2$ / $256$ / $8$ \\
Decoder & Experts / routing & $4$ / learned soft top-$2$ \\
 & Expert MLP widths & $128\rightarrow256\rightarrow128$ \\
\midrule
Both stages & Optimizer / learning rate / weight decay & AdamW / $10^{-4}$ / $10^{-4}$ \\
 & Batch size & $64$ reference-adapter samples \\
Stage 1 & Training duration & $300$ epochs \\
 & Block / cosine / KL coefficients & $1.0$ / $0.1$ / $10^{-4}$ \\
 & Functional / balance coefficients & $1.0$ / $0.01$ \\
Stage 2 & DiT layers / hidden width / heads & $4$ / $256$ / $8$ \\
 & Training diffusion horizon / schedule / prediction & $1000$ / cosine / $\epsilon$ \\
 & Training duration & $10{,}000$ optimizer updates \\
 & Diffusion / latent / weight coefficients & $1.0$ / $0.1$ / $0.1$ \\
 & Functional coefficient & $1.0$ \\
 & Auxiliary reconstruction/function losses & $\bar\alpha_t>0.5$ (SNR $>1$) \\
Training & Gradient clipping threshold / EMA decay & $1.0$ / $0.9999$ \\
\midrule
Source bank & Size / eligibility & $311$ per backbone / all eligible \\
 & Sampling & Dataset-balanced \\
Inference & DDIM~\citep{song2020ddim} steps / $\eta_{\mathrm{DDIM}}$ & $50$ / $0$ \\
 & Calibration per subject & $64$ unlabeled windows \\
 & Sleep window duration & $30$ s \\
Evaluation & Matched seeds & $5$ \\
\bottomrule
\end{tabular}
\end{table*}

\section{Experiments}
\label{sec:experiments}

\subsection{Experimental Setup}
\label{sec:exp_setup}

\paragraph{Datasets and Tasks}
We evaluate ZeroMAG on a multi-dataset benchmark spanning sleep staging,
emotion recognition, cognitive load, motor imagery, and
freezing-of-gait detection. The recordings cover EOG, EMG, ECG, eye movement,
acceleration, and skin conductance, including settings in which modality
combinations or modality types are excluded from generator training.
Table~\ref{tab:main} reports results on six target datasets that are
disjoint from the source training datasets; FoG is also a task-holdout
target with no source adapter for its task. Task coverage appears in Table~\ref{tab:dataset_tasks};
Table~\ref{tab:extra_targets} reports the remaining three targets.
Appendix~C gives extended analyses.

\paragraph{Backbones and Baselines}
We use three frozen EEG foundation models: CBraMod~\citep{cbramod},
CSBrain~\citep{csbrain}, and CodeBrain~\citep{codebrain}.
Each backbone has a separately trained generator. ZeroMAG and the
target-label-free baselines use the same frozen EEG encoder and prediction
head. These baselines include EEG-only inference, mean and nearest source
weights, and Direct MLP. A separate supervised multimodal fine-tuning
experiment uses target training labels to measure the gap to supervised
adaptation on the same query split; it is not part of the ZeroMAG pipeline.
PhysioOmni~\citep{physioomni}, SleepFM~\citep{sleepfm2024}, and
Brant-X~\citep{brantx} serve as cross-paradigm references.

\subsection{Data Splits and Evaluation Protocol}
\label{sec:dataset_protocol}
The adapter-bank and generator source datasets are ISRUC-S1~\citep{isruc},
Sleep-EDF Expanded~\citep{sleepedf}, SHHS~\citep{shhs}, DEAP~\citep{deap},
ASCERTAIN~\citep{ascertain}, SEED-VIG~\citep{seedvig}, WAUC~\citep{wauc},
and OpenBMI~\citep{openbmi}.
The target datasets are ISRUC-S3~\citep{isruc}, HMC~\citep{hmc},
MASS-SS3 (reported as MASS)~\citep{mass}, SEED-V~\citep{seedv},
EEGMAT~\citep{eegmat}, Multimodal FoG~\citep{fog}, CAP~\citep{cap},
DREAMER~\citep{dreamer}, and BCI-IV-2a (reported as BCI-IV)~\citep{bciiv2a}.
Let $\mathcal D_{\mathrm{src}}$ and $\mathcal D_{\mathrm{tgt}}$ denote
these two disjoint sets:
\begin{equation}
\mathcal D_{\mathrm{src}}\cap\mathcal D_{\mathrm{tgt}}=\varnothing,
\qquad
|\mathcal D_{\mathrm{tgt}}|=9.
\end{equation}
All target datasets are excluded from encoder/head training, source
reference-adapter fitting, both generator-training stages, and model
selection in the ZeroMAG pipeline. Reference-adapter supervision comes from source
training data; target recordings are used only for unlabeled conditioning
and evaluation.

For a source recording, $R^{\mathrm{cond}}\subseteq R^{\mathrm{fit}}$ is read without its labels. The functional subset is separate:
\begin{equation}
R^{\mathrm{fun}}\cap R^{\mathrm{fit}}=\varnothing,
\qquad R^{\mathrm{fun}}\cap R^{\mathrm{cond}}=\varnothing.
\end{equation}
The labels associated with $R^{\mathrm{fit}}$ train the reference adapter. Functional supervision compares predictions on $R^{\mathrm{fun}}$ without reading its labels. These distinct roles prevent the condition encoder's input from being confused with the reference adapter's supervised fitting data.

For each target subject, calibration and query subsets are disjoint:
\begin{equation}
R_{\mathrm{tgt}}^{\mathrm{cond}}\cap R_{\mathrm{tgt}}^{\mathrm{qry}}=\varnothing.
\end{equation}
The condition is constructed from calibration signals and the supported task. The generated adapter is evaluated on query inputs without target-side gradient updates. Query labels serve evaluation rather than adapter fitting or condition construction.

\subsection{Metrics and Implementation}
\label{sec:implementation_main}
We report balanced accuracy (B-ACC) as mean and standard deviation over five
matched seeds. The six-target summary gives equal weight to the 18
backbone--dataset means in Table~\ref{tab:main}, whereas the CBraMod ablation
averages all nine targets in an independent experimental batch. Seed
matching is within each comparison, not between these batches.
Table~\ref{tab:implementation} lists the training and inference settings.
Batch size counts reference-adapter samples, not subjects or signal windows.
The source bank contains 311 eligible adapters per backbone and uses
dataset-balanced sampling. Each target subject supplies 64 unlabeled
calibration windows. The mean/SD summaries do not establish paired
significance or statistical equivalence; further metric and aggregation
details are in Appendix~H.

\subsection{Performance Comparison with Baselines}
\label{sec:exp_main}

\begin{table*}[t]
\centering
\caption{
Balanced accuracy (\%, mean$\pm$std over five seeds) on six held-out
target datasets. Within each EEG-backbone group, ZeroMAG and the label-free
baselines use the same frozen encoder and prediction head, without target
labels or target-side optimization. Supervised multimodal fine-tuning is
a separate reference using target training labels and all available
modalities. Bold entries indicate the best target-label-free result in
each group.
}
\label{tab:main}

\small
\setlength{\tabcolsep}{4.5pt}
\renewcommand{\arraystretch}{1.12}

\begin{tabular}{@{}lcccccc@{}}
\toprule
\textbf{Method}
& \textbf{ISRUC-S3}
& \textbf{HMC}
& \textbf{MASS}
& \textbf{SEED-V}
& \textbf{EEGMAT}
& \textbf{FoG} \\
\midrule

\multicolumn{7}{@{}l}{\textit{CBraMod}} \\
\addlinespace[1pt]

\hspace{0.8em}EEG-only
& \ms{75.83}{0.76}
& \ms{72.78}{0.50}
& \ms{74.29}{0.58}
& \ms{40.62}{1.24}
& \ms{73.38}{2.48}
& \ms{76.61}{1.35} \\

\hspace{0.8em}Supervised tuning
& \ms{84.04}{0.88}
& \ms{80.81}{0.41}
& \ms{82.41}{0.68}
& \ms{45.44}{1.58}
& \ms{81.46}{1.93}
& \ms{85.02}{1.21} \\

\hspace{0.8em}Mean source weights
& \ms{74.31}{0.82}
& \ms{72.93}{0.49}
& \ms{75.91}{0.63}
& \ms{39.83}{1.13}
& \ms{73.52}{0.92}
& \ms{76.73}{0.78} \\

\hspace{0.8em}Nearest source weights
& \ms{76.63}{0.61}
& \ms{73.18}{0.52}
& \ms{76.35}{0.51}
& \ms{40.68}{1.37}
& \ms{74.63}{0.81}
& \ms{77.52}{0.73} \\

\hspace{0.8em}Direct MLP
& \ms{76.89}{0.73}
& \ms{73.43}{0.55}
& \ms{77.48}{0.65}
& \ms{40.99}{1.05}
& \ms{75.21}{1.02}
& \ms{78.03}{0.62} \\

\rowcolor{orange!7}
\hspace{0.8em}\textbf{ZeroMAG}
& \textbf{\ms{83.59}{0.59}}
& \textbf{\ms{80.29}{0.42}}
& \textbf{\ms{82.03}{0.52}}
& \textbf{\ms{44.83}{1.22}}
& \textbf{\ms{80.99}{0.81}}
& \textbf{\ms{84.47}{0.72}} \\

\addlinespace[5pt]

\multicolumn{7}{@{}l}{\textit{CSBrain}} \\
\addlinespace[1pt]

\hspace{0.8em}EEG-only
& \ms{75.94}{0.93}
& \ms{73.31}{0.48}
& \ms{75.89}{0.54}
& \ms{41.38}{1.32}
& \ms{75.38}{2.67}
& \ms{75.96}{1.52} \\

\hspace{0.8em}Supervised tuning
& \ms{84.10}{1.43}
& \ms{81.39}{0.39}
& \ms{84.15}{0.62}
& \ms{46.32}{1.17}
& \ms{83.57}{1.43}
& \ms{85.13}{1.31} \\

\hspace{0.8em}Mean source weights
& \ms{75.42}{0.62}
& \ms{73.51}{0.45}
& \ms{77.72}{0.58}
& \ms{41.75}{0.62}
& \ms{75.91}{0.71}
& \ms{77.08}{0.65} \\

\hspace{0.8em}Nearest source weights
& \ms{76.74}{0.55}
& \ms{73.82}{0.52}
& \ms{77.98}{0.63}
& \ms{42.28}{0.55}
& \ms{75.83}{0.66}
& \ms{78.73}{0.59} \\

\hspace{0.8em}Direct MLP
& \ms{79.18}{0.71}
& \ms{74.21}{0.48}
& \ms{80.45}{0.72}
& \ms{42.69}{0.48}
& \ms{78.12}{0.58}
& \ms{80.02}{0.52} \\

\rowcolor{orange!7}
\hspace{0.8em}\textbf{ZeroMAG}
& \textbf{\ms{83.67}{0.61}}
& \textbf{\ms{80.83}{0.52}}
& \textbf{\ms{83.74}{0.83}}
& \textbf{\ms{45.74}{0.44}}
& \textbf{\ms{83.08}{0.71}}
& \textbf{\ms{84.60}{0.65}} \\

\addlinespace[5pt]

\multicolumn{7}{@{}l}{\textit{CodeBrain}} \\
\addlinespace[1pt]

\hspace{0.8em}EEG-only
& \ms{76.57}{0.73}
& \ms{72.95}{0.49}
& \ms{77.24}{0.51}
& \ms{41.52}{1.46}
& \ms{76.62}{2.83}
& \ms{76.77}{1.26} \\

\hspace{0.8em}Supervised tuning
& \ms{84.92}{1.61}
& \ms{80.93}{0.38}
& \ms{85.67}{0.58}
& \ms{46.39}{1.34}
& \ms{85.02}{1.63}
& \ms{85.25}{1.33} \\

\hspace{0.8em}Mean source weights
& \ms{77.02}{0.68}
& \ms{73.28}{0.42}
& \ms{79.12}{0.72}
& \ms{41.88}{0.58}
& \ms{77.22}{0.66}
& \ms{77.85}{0.71} \\

\hspace{0.8em}Nearest source weights
& \ms{77.22}{0.62}
& \ms{73.65}{0.48}
& \ms{80.21}{0.68}
& \ms{42.35}{0.52}
& \ms{77.98}{0.59}
& \ms{79.22}{0.65} \\

\hspace{0.8em}Direct MLP
& \ms{79.92}{0.55}
& \ms{73.92}{0.51}
& \ms{81.65}{0.61}
& \ms{42.87}{0.45}
& \ms{79.28}{0.52}
& \ms{80.55}{0.58} \\

\rowcolor{orange!7}
\hspace{0.8em}\textbf{ZeroMAG}
& \textbf{\ms{84.44}{0.77}}
& \textbf{\ms{80.43}{0.48}}
& \textbf{\ms{85.25}{0.72}}
& \textbf{\ms{45.84}{0.55}}
& \textbf{\ms{84.51}{0.63}}
& \textbf{\ms{84.65}{0.58}} \\

\addlinespace[5pt]
\midrule
\addlinespace[2pt]

\multicolumn{7}{@{}l}{\textit{Multimodal physiological models}} \\
\addlinespace[1pt]

\hspace{0.8em}PhysioOmni
& \ms{84.62}{0.63}
& \ms{81.05}{0.42}
& \ms{85.50}{0.65}
& \ms{46.04}{1.32}
& \ms{84.81}{2.13}
& \ms{84.93}{0.53} \\

\hspace{0.8em}SleepFM
& \ms{83.27}{0.61}
& \ms{80.01}{0.47}
& \ms{81.61}{0.52}
& \ms{44.48}{0.58}
& \ms{80.54}{0.72}
& \ms{84.09}{0.68} \\

\hspace{0.8em}Brant-X
& \ms{85.19}{0.54}
& \ms{81.51}{0.46}
& \ms{86.07}{0.59}
& \ms{46.49}{0.63}
& \ms{85.31}{0.61}
& \ms{85.37}{0.72} \\

\bottomrule
\end{tabular}
\end{table*}
\begin{table}[!t]\centering\footnotesize
\caption{B-ACC (\%, mean$\pm$std over five seeds) on CAP, DREAMER, and BCI-IV. Protocol and layout as in Table~\ref{tab:main}.}\label{tab:extra_targets}
\setlength{\tabcolsep}{2.5pt}\renewcommand{\arraystretch}{1.12}
\begin{tabular}{@{}lccc@{}}\toprule
\textbf{Method} & \textbf{CAP} & \textbf{DREAMER} & \textbf{BCI-IV} \\\midrule
\multicolumn{4}{@{}l}{\textit{CBraMod}} \\
\addlinespace[1pt]
\hspace{0.8em}EEG-only & \ms{53.86}{0.83} & \ms{43.91}{1.04} & \ms{37.82}{1.31} \\
\hspace{0.8em}Supervised tuning & \ms{68.16}{0.64} & \ms{60.20}{0.98} & \ms{51.11}{0.89} \\
\hspace{0.8em}Mean source weights & \ms{53.68}{0.97} & \ms{43.45}{1.35} & \ms{37.54}{1.70} \\
\hspace{0.8em}Nearest source weights & \ms{54.78}{0.55} & \ms{45.08}{1.23} & \ms{39.26}{0.91} \\
\hspace{0.8em}Direct MLP & \ms{55.93}{0.65} & \ms{46.05}{0.74} & \ms{41.00}{1.61} \\
\rowcolor{orange!7}
\hspace{0.8em}\textbf{ZeroMAG} & \textbf{\ms{66.11}{0.69}} & \textbf{\ms{58.07}{0.98}} & \textbf{\ms{49.69}{1.72}} \\
\addlinespace[4pt]
\multicolumn{4}{@{}l}{\textit{CSBrain}} \\
\addlinespace[1pt]
\hspace{0.8em}EEG-only & \ms{54.23}{0.79} & \ms{44.38}{0.98} & \ms{38.57}{1.26} \\
\hspace{0.8em}Supervised tuning & \ms{66.30}{0.97} & \ms{59.77}{1.05} & \ms{52.62}{1.06} \\
\hspace{0.8em}Mean source weights & \ms{54.67}{1.02} & \ms{43.71}{1.07} & \ms{38.41}{1.21} \\
\hspace{0.8em}Nearest source weights & \ms{55.59}{1.06} & \ms{45.92}{0.94} & \ms{39.54}{1.50} \\
\hspace{0.8em}Direct MLP & \ms{56.91}{1.07} & \ms{46.85}{1.34} & \ms{40.29}{1.77} \\
\rowcolor{orange!7}
\hspace{0.8em}\textbf{ZeroMAG} & \textbf{\ms{64.49}{0.56}} & \textbf{\ms{58.10}{1.05}} & \textbf{\ms{51.24}{1.11}} \\
\addlinespace[4pt]
\multicolumn{4}{@{}l}{\textit{CodeBrain}} \\
\addlinespace[1pt]
\hspace{0.8em}EEG-only & \ms{53.12}{0.91} & \ms{43.46}{1.11} & \ms{37.24}{1.38} \\
\hspace{0.8em}Supervised tuning & \ms{67.50}{0.59} & \ms{55.46}{0.96} & \ms{50.23}{1.27} \\
\hspace{0.8em}Mean source weights & \ms{52.50}{0.70} & \ms{43.40}{0.97} & \ms{37.02}{0.93} \\
\hspace{0.8em}Nearest source weights & \ms{53.77}{0.55} & \ms{44.37}{0.78} & \ms{38.48}{1.33} \\
\hspace{0.8em}Direct MLP & \ms{54.19}{1.03} & \ms{45.14}{0.95} & \ms{38.87}{1.76} \\
\rowcolor{orange!7}
\hspace{0.8em}\textbf{ZeroMAG} & \textbf{\ms{65.60}{0.74}} & \textbf{\ms{54.51}{1.14}} & \textbf{\ms{48.09}{1.71}} \\
\bottomrule
\end{tabular}\end{table}
\begin{figure*}[!t]\centering
\includegraphics[width=.62\textwidth]{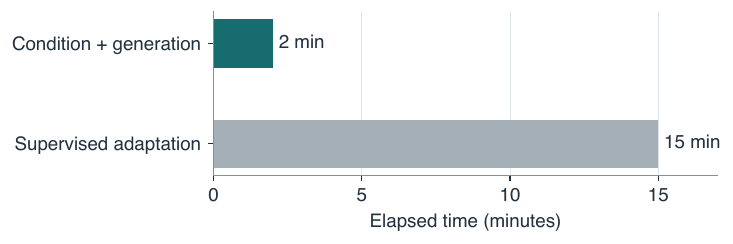}
\caption{Target-side wall-clock time measured programmatically: condition construction plus adapter generation versus supervised adaptation, under the same hardware and protocol. The 7.5$\times$ ratio uses the reported 2- and 15-minute measurements. No timing variance or component-wise timing breakdown is inferred.}\label{fig:app_timing}
\end{figure*}
\begin{figure*}[!t]\centering
\includegraphics[width=.80\textwidth]{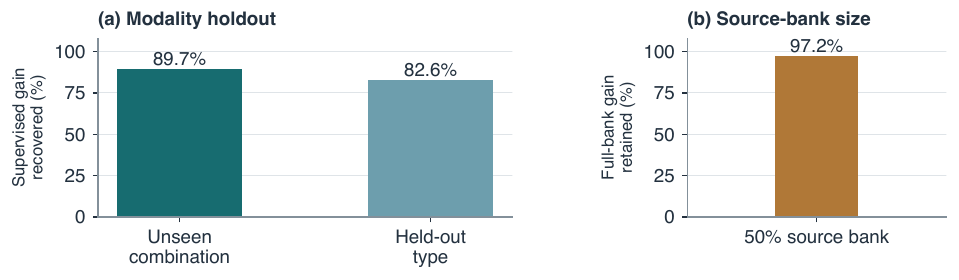}
\caption{Reported generalization and source-bank summaries from the main text. (a) Recovery of supervised gain for unseen combinations and held-out modality types. (b) Retention of the full-bank multimodal gain when using half the source bank. The panels have different reference denominators; no seed-level intervals are inferred.}\label{fig:app_generalization}
\end{figure*}

\paragraph{Comparison with label-free and supervised adaptation}
ZeroMAG ranks first among target-label-free methods in all 18
backbone--dataset evaluations, improving B-ACC on average by 7.22 points
over EEG-only inference, 6.78 over mean source weights, 6.00 over nearest
source weights, and 4.89 over Direct MLP, the strongest competing
generator. Despite using neither target labels nor target-side
optimization, it remains only 0.38--0.61 points below full-modality
supervised tuning in every setting of Table~\ref{tab:main} (0.50 on average), recovering 93.50\%
of the supervised gain over EEG-only inference.

\paragraph{Backbone and cross-paradigm generalization}
The gain over EEG-only inference is stable across CBraMod, CSBrain, and
CodeBrain at 7.12, 7.30, and 7.24 points; the range is approximately 0.19 points.
With CodeBrain, ZeroMAG exceeds SleepFM on all six displayed targets by
1.85 points on average, trails PhysioOmni by only 0.31 points, and
trails Brant-X by 0.80 points on average.

\paragraph{Additional held-out targets}
Table~\ref{tab:extra_targets} extends the comparison to CAP, DREAMER,
and BCI-IV-2a, completing the nine-target evaluation. It uses the same frozen
interfaces and reports the original five-seed means and standard deviations.
The headline six-target averages above retain their original scope. On these three targets, the supervised-minus-ZeroMAG gap is larger, ranging from 0.95 to 2.14 points.

\subsection{Ablation Study}
\label{sec:exp_ablation}

We ablate the condition, generator, adapter, and functional supervision.
Within the ablation batch, we keep the source adapter bank, source
training splits, frozen backbone, prediction heads, and evaluation
protocol fixed. Table~\ref{tab:ablation} reports CBraMod results averaged
over all nine held-out target datasets. These experiments were conducted
in a separate batch from the main performance comparisons in
Tables~\ref{tab:main} and~\ref{tab:extra_targets}. Each ablated variant
is compared with Full ZeroMAG from the same ablation batch.

\definecolor{ablationblue}{RGB}{225,239,249}

\begin{table}[!t]
\centering
\caption{
Ablation results with CBraMod, averaged over all nine held-out
datasets (\% B-ACC, mean$\pm$std over five seeds).
Ablated results are shaded blue, and $\Delta$ denotes the decrease from
Full ZeroMAG within the same ablation batch. This batch is independent of
the main performance comparisons; see Appendix~G.1.
}
\label{tab:ablation}

\footnotesize
\setlength{\tabcolsep}{2.6pt}
\renewcommand{\arraystretch}{1.10}

\begin{tabular}{@{}llcc@{}}
\toprule
\textbf{Component}
& \textbf{Variant}
& \textbf{B-ACC $\uparrow$}
& \textbf{$\Delta$ $\downarrow$} \\
\midrule

Condition
& Task + modality only
& \cellcolor{ablationblue}\ms{62.32}{0.41}
& \cellcolor{ablationblue}\ms{11.09}{0.72} \\

& Shuffled subject condition
& \cellcolor{ablationblue}\ms{64.17}{0.53}
& \cellcolor{ablationblue}\ms{9.24}{0.98} \\

\addlinespace[3pt]

Generator
& Direct weight MLP
& \cellcolor{ablationblue}\ms{62.34}{0.67}
& \cellcolor{ablationblue}\ms{11.07}{0.88} \\

& Modular latent MLP
& \cellcolor{ablationblue}\ms{65.23}{0.62}
& \cellcolor{ablationblue}\ms{8.18}{0.83} \\

& Modular latent DiT (w/o MoE)
& \cellcolor{ablationblue}\ms{68.34}{0.21}
& \cellcolor{ablationblue}\ms{5.07}{0.71} \\

\addlinespace[3pt]

Adapter
& Shared modality encoder
& \cellcolor{ablationblue}\ms{69.25}{0.33}
& \cellcolor{ablationblue}\ms{4.16}{0.81} \\

& Temporal branch only
& \cellcolor{ablationblue}\ms{65.29}{0.72}
& \cellcolor{ablationblue}\ms{8.12}{1.03} \\

& Mean fusion
& \cellcolor{ablationblue}\ms{66.83}{0.32}
& \cellcolor{ablationblue}\ms{6.58}{0.71} \\

\addlinespace[3pt]

Functional
& w/o Stage 1 functional
& \cellcolor{ablationblue}\ms{72.26}{0.76}
& \cellcolor{ablationblue}1.15 \\

& w/o Stage 2 functional
& \cellcolor{ablationblue}\ms{71.83}{0.79}
& \cellcolor{ablationblue}1.58 \\

& w/o both
& \cellcolor{ablationblue}\ms{70.54}{0.86}
& \cellcolor{ablationblue}2.87 \\

\addlinespace[3pt]
\midrule

\multicolumn{2}{@{}l}{\textbf{Full ZeroMAG}}
& \textbf{\ms{73.41}{0.71}}
& --- \\

\bottomrule
\end{tabular}
\end{table}
\paragraph{Condition}
Removing the subject factor and retaining only task and modality
information lowers B-ACC by 11.09 points. Shuffling subject conditions produces a similarly
large 9.24-point drop, although the representations retain the same
marginal distribution and dimensionality. The condition must therefore
encode subject-specific information aligned with the target recording.

\paragraph{Generator}
Direct weight regression decreases B-ACC by 11.07 points. The modular
latent space raises B-ACC from 62.34 to 65.23, DiT to 68.34, and the MoE
decoder to 73.41.

\paragraph{Adapter}
Sharing the modality encoder reduces B-ACC by 4.16 points, removing the
spectral branch by 8.12 points, and replacing learned coordination with
mean fusion by 6.58 points, supporting modality-specific encoding,
temporal--spectral processing, and learned fusion.

\subsection{Functional Supervision Analysis}
\label{sec:functional_main}
Within the same experimental batch as Table~\ref{tab:ablation}, we evaluate four functional-loss settings using CBraMod on the same nine held-out target datasets. The variants use five matched random seeds, the same data splits and reference-adapter banks, and matched training budgets, initialization rules, and evaluation protocols. B-ACC is reported as mean $\pm$ standard deviation over the five seeds.

\begin{table}[!t]
\centering\footnotesize
\setlength{\tabcolsep}{2.2pt}
\caption{Functional-loss ablation. Parameter-error and KL columns are normalized to their respective Full values; lower is better. Reconstruction error evaluates the Stage 1 output against the reference adapter. Generated-weight error evaluates the final generated adapter against the same reference. KL evaluates the final generated adapter.}
\label{tab:functional_ablation}
\begin{tabular}{lrrrr}
\toprule
Setting & B-ACC (\%) & \shortstack{Stage 1\\recon. error} & \shortstack{Generated\\weight error} & \shortstack{Prediction\\KL}\\
\midrule
Full & $73.41\pm0.71$ & 1.00 & 1.00 & 1.00\\
w/o Stage 1 functional & $72.26\pm0.76$ & 0.94 & 1.05 & 1.31\\
w/o Stage 2 functional & $71.83\pm0.79$ & 1.00 & 1.08 & 1.48\\
w/o both functional & $70.54\pm0.86$ & 0.93 & 1.14 & 1.82\\
\bottomrule
\end{tabular}
\end{table}

\paragraph{Evaluation}
Reconstruction and generated-weight errors use the same parameter-distance definition, comparing $\widetilde\omega$ and $\widehat\omega$, respectively, with the supervised reference $\omega^*$. The prediction discrepancy is $D_{\mathrm{KL}}(p_{\omega^*}\|p_{\widehat\omega})$, evaluated on a fixed independent evaluation subset and averaged over samples, targets, and seeds. Full ZeroMAG is the comparison baseline in Table~\ref{tab:functional_ablation}: each error is divided by the corresponding error of Full ZeroMAG. Its normalized values are therefore 1.00 by definition. The normalization baseline is distinct from the reference adapter $\omega^*$ used in the underlying error definitions; the three normalized metrics need not have equal absolute values.

\paragraph{Results}
Removing Stage 1 functional supervision reduces reconstruction error by 6\%, yet increases final generated-weight error by 5\% and prediction KL by 31\%; B-ACC falls by 1.15 percentage points. Thus, better reconstruction of reference parameters does not necessarily yield better generated-adapter predictions. Removing Stage 2 functional supervision leaves Stage 1 reconstruction error unchanged, while generated-weight error increases by 8\%, prediction KL by 48\%, and B-ACC decreases by 1.58 points. Removing both losses gives the largest observed B-ACC decrease, 2.87 points, and raises prediction KL by 82\%. These results support functional supervision during adapter representation learning and conditional generation.

\subsection{Generalization and Efficiency}
\label{sec:exp_generalization}

\paragraph{Unseen modalities}
On unseen combinations of known modalities and modality types in
$\mathcal{M}$ excluded from generator training, ZeroMAG recovers 89.7\% and 82.6\% of the gain
achieved by supervised multimodal adaptation, respectively. It
outperforms mean source transfer, nearest-source retrieval, and direct
weight regression in both settings.

\paragraph{Data and target-side efficiency}
Performance saturates at approximately 64 unlabeled calibration
windows. Moreover, using only 50\% of the source adapter bank retains
97.2\% of the full multimodal gain. ZeroMAG constructs the condition and generates an adapter in 2 minutes,
compared with 15 minutes for supervised target adaptation under the
same hardware and protocol ($7.5\times$ faster); offline training cost is
amortized across targets (Appendix~F).

Figure~\ref{fig:app_timing} compares measured target adaptation time;
it excludes source-bank construction and offline training, and does not
imply a speedup for every subsequent query prediction.
The percentages in Fig.~\ref{fig:app_generalization} have different
reference denominators: supervised gain for modality holdout and full-bank
gain for source-bank reduction. They are not B-ACC percentage-point changes.

\section{Conclusion}
\label{sec:conclusion}

We introduced ZeroMAG, a zero-shot framework that adds multimodal fusion
capability to frozen EFMs through generated plug-in adapters. ZeroMAG fuses an
available combination of supported companion physiological signals around an
invariant EEG anchor, integrates recording-derived modality and subject
information with supplied task context into a single generation condition, and generates the fusion
parameters in a function-constrained latent space learned from source
adapters. It consistently improves over target-label-free baselines and
approaches supervised multimodal adaptation, requiring no target labels or
target-side optimization for the added adapter and no repeated multimodal
pretraining. More broadly, the results indicate that the fusion module of a
multimodal system can be inferred from unlabeled target data instead of being
trained separately for each target, a setting that arises whenever a strong
unimodal backbone must be combined with a variable set of auxiliary sensors.

ZeroMAG has several limitations. It relies on a supplied fine-tuned EFM and
adequate source-bank coverage, and it cannot incorporate modality types
outside $\mathcal{M}$ without retraining. Enlarging the source adapter bank to
cover more modality types and tasks is a natural direction for extending the
range of configurations that can be fused without target-side training.


\section*{Declaration of competing interest}
The authors declare that they have no known competing financial interests or
personal relationships that could have appeared to influence the work reported
in this paper.

\section*{Declaration of generative AI and AI-assisted technologies in the manuscript preparation process}
During the preparation of this work, the authors used ChatGPT, Codex, and
Claude for assistance with language editing. Codex was also used for LaTeX
revision, reference checking, and manuscript consistency checks. All figures
were drawn by the authors without generative AI tools.
These tools were not used in this revision to run experiments or generate or
alter experimental measurements. After using these tools, the authors
reviewed and edited the content as needed and take full responsibility for
the content of the published article.

\section*{Funding}
This research did not receive any specific grant from funding agencies in the public, commercial, or not-for-profit sectors.

\section*{Data and code availability}
All datasets used in this study are available from their original providers
and are listed in Appendix~C; some require a data-use
agreement with the provider. Implementation details and hyperparameters are
listed in Table~\ref{tab:implementation}, and split details are given in
Section~\ref{sec:dataset_protocol}. An implementation snapshot is available at
\url{https://github.com/Yubo-CY/ZeroMAG}.

\section*{Appendices}
The appendices following the references contain notation, adapter details,
expanded result analyses, training and inference algorithms, objective
derivations, efficiency analysis, and evaluation details.

\bibliographystyle{elsarticle-num-names}
\bibliography{refs}

\clearpage
\onecolumn
\appendix
\renewcommand{\thesection}{\Alph{section}}
\renewcommand{\thesubsection}{\thesection.\arabic{subsection}}
\counterwithin{equation}{section}
\counterwithin{figure}{section}
\counterwithin{table}{section}
\counterwithin{algorithm}{section}

\renewcommand{\theequation}{\thesection.\arabic{equation}}
\renewcommand{\thefigure}{\thesection.\arabic{figure}}
\renewcommand{\thetable}{\thesection.\arabic{table}}
\renewcommand{\theHequation}{app.\thesection.\arabic{equation}}
\renewcommand{\theHfigure}{app.\thesection.\arabic{figure}}
\renewcommand{\theHtable}{app.\thesection.\arabic{table}}

\section*{Appendix Contents}
\begingroup\small
\newcommand{\appcontentsline}[3]{\noindent\hyperref[#2]{#1. #3}\dotfill\pageref{#2}\par}
\noindent\begin{minipage}[t]{.48\linewidth}
\appcontentsline{A}{sec:app_notation}{Notation and Model Scope}
\appcontentsline{B}{sec:app_components}{Adapter and Condition}
\appcontentsline{C}{sec:app_datasets}{Datasets and Additional Results}
\appcontentsline{D}{sec:app_procedure}{Training and Deployment}
\appcontentsline{E}{sec:app_derivations}{Objective Analysis}
\end{minipage}\hfill
\begin{minipage}[t]{.48\linewidth}
\appcontentsline{F}{sec:app_efficiency}{Efficiency}
\appcontentsline{G}{sec:app_ablations}{Component Ablations}
\appcontentsline{H}{sec:app_protocol}{Implementation and Evaluation Protocol}
\appcontentsline{I}{sec:app_robustness}{Generalization and Data Efficiency}
\appcontentsline{J}{sec:app_limitations}{Comparison with Related Methods}
\end{minipage}
\endgroup
\medskip

\section{Notation and Model Scope}
\label{sec:app_notation}
\subsection{EEG representations}
The adapter interface is defined by the frozen EEG representation and compatible prediction head. The three backbone instantiations are described in Section~3.3.

\subsection{Frozen interface}\label{sec:app_frozen_models}
The frozen prediction-head interface and separate backbone instantiations are specified in Section~3.3. The notation below applies to the selected instantiation.

\subsection{Signals and adapter parameters}
Tables~\ref{tab:notation_signals} and~\ref{tab:notation_latents} collect the notation used below. As in the main text, $K$ denotes the number of classes and $C$ the fused condition.

\begingroup\small
\begin{longtable}{p{.22\linewidth}p{.70\linewidth}}
\caption{Signal and adapter notation. Dimensions are symbolic here; numerical settings are listed in Table~2.}\label{tab:notation_signals}\\
\toprule Symbol & Meaning\\\midrule\endfirsthead
\toprule Symbol & Meaning\\\midrule\endhead
$X^{\mathrm{EEG}}$ & EEG input to the frozen encoder.\\
$\mathcal M$ & Complete set of supported companion modalities.\\
$A\subseteq\mathcal M$ & Companion modalities available in a particular recording.\\
$X^A$ & Collection $\{X^m:m\in A\}$ of available companion inputs.\\
$R=\{r_i\}$ & Recording samples, with $r_i=(X_i^{\mathrm{EEG}},X_i^A)$.\\
$R^{\mathrm{fit}}$ & Source samples whose labels are used to fit a reference adapter.\\
$R^{\mathrm{cond}}$ & Unlabeled samples used to construct the condition.\\
$R^{\mathrm{fun}}$ & Source samples used for functional supervision, disjoint from fitting and conditioning.\\
$R^{\mathrm{qry}}$ & Target query samples, disjoint from target conditioning.\\
$\tau$ & Supported prediction task associated with the frozen head.\\
$f,h$ & Frozen EEG encoder and prediction head.\\
$Z=f(X^{\mathrm{EEG}})$ & EEG representation before the adapter.\\
$a(Z,X^A;\omega)$ & Multimodal adapter, with parameter vector $\omega\in\mathbb R^{P(A)}$.\\
$Z_A,L^A(\omega)$ & Adapted representation and logits $h(Z_A)$.\\
$\omega^*,\widetilde\omega,\widehat\omega$ & Fitted reference, reconstructed reference, and generated adapter.\\
$\omega^{\mathrm{sh}}$ & Parameters of the shared coordinator.\\
$\omega^m_{\mathrm{temp}},\omega^m_{\mathrm{spec}},\omega^m_{\mathrm{proj}}$ & Temporal, spectral, and projection parameters for modality $m$.\\
$B,p_b,P(A)$ & Active block count $B=3|A|+1$, block size, and configuration-specific parameter count.\\
\bottomrule
\end{longtable}
\endgroup

\subsection{Latents and conditioning}
The pre-contextual adapter token is denoted $U_{b,\mathrm{pre}}$, following the main text. It is distinct from the signal features $V_i^m$ and $V^m$ used for conditioning. The deterministic latent $U_0$ is obtained by substituting posterior means before the cross-block transformer; it is not a new encoder architecture.

\begin{longtable}{p{.22\linewidth}p{.70\linewidth}}
\caption{Latent, condition, and optimization notation.}\label{tab:notation_latents}\\
\toprule Symbol & Meaning\\\midrule\endfirsthead
\toprule Symbol & Meaning\\\midrule\endhead
$e(\cdot;\eta)$ & Stochastic block encoder, including the cross-block transformer.\\
$\bar e(\cdot;\eta)$ & Deterministic encoder path using posterior means.\\
$\mu_b,\ell_b$ & Mean and log variance of the Gaussian for block $b$; $\ell_b$ is introduced here for the posterior expansion.\\
$U_{b,\mathrm{pre}}$ & Sampled block token before contextualization.\\
$E_b,\mathrm{CBT}$ & Block identity embedding and cross-block transformer.\\
$U,U_0,U_t$ & Contextualized sample, canonical latent, and noise-corrupted canonical latent.\\
$\widehat U_0$ & Clean-latent estimate computed from a noisy latent and predicted noise.\\
$d(\cdot;\psi)$ & MoE decoder mapping latent tokens to adapter weights.\\
$\pi_{b,j},r_j$ & Routing weight and reconstruction expert for expert $j$.\\
$V_i^m,V^m$ & Sample-level and recording-level condition features for modality $m$.\\
$e^m_{\mathrm{mod}},q_{\mathrm{sub}}$ & Modality embedding and task-independent attention query.\\
$F_{\mathrm{mod}},F_{\mathrm{sub}},E_{\mathrm{task}}$ & Set pooling, attention pooling, and task embedding.\\
$C_{\mathrm{mod}},C_{\mathrm{sub}},C_\tau$ & Modality, subject, and task factors.\\
$C,E_{\mathrm{cond}}(\cdot;\xi),F_{\mathrm{cond}}$ & Fused condition, complete condition encoder, and factor-fusion function.\\
$\epsilon,\epsilon_\theta,s_\theta$ & Gaussian noise, conditional noise predictor, and 50-step deterministic DDIM sampler using $\epsilon_\theta$.\\
$t,\bar\alpha_t$ & Diffusion timestep and cumulative signal coefficient.\\
$\delta_{\mathrm{fun}}$ & Reference-to-student predictive KL averaged on $R^{\mathrm{fun}}$.\\
$\mathcal L_{\mathrm{AE}},\mathcal L_{\mathrm{gen}}$ & Stage 1 and Stage 2 objectives.\\
\bottomrule
\end{longtable}

\subsection{Parameter ownership}
The learning stages update different parameter sets. Table~\ref{tab:parameter_ownership} separates parameter updates from differentiable use. A frozen decoder remains part of the computational graph for a loss on its output, even though the optimizer does not change its parameters.

\begin{table}[htbp]\centering\small
\caption{Parameter updates in the ZeroMAG pipeline. ``Fixed'' describes parameter values, not a zero input Jacobian.}\label{tab:parameter_ownership}
\begin{tabular}{lcccc}
\toprule Component & Reference fitting & Stage 1 & Stage 2 & Deployment\\\midrule
EEG encoder and head & Fixed & Fixed & Fixed & Fixed\\
Reference adapter $\omega^*$ & Updated & Fixed & Fixed & Not required\\
Encoder $\eta$ & Not used & Updated & Fixed & Not required\\
Decoder $\psi$ & Not used & Updated & Fixed & Fixed\\
Condition encoder $\xi$ & Not used & Not used & Updated & Fixed\\
Denoiser $\theta$ & Not used & Not used & Updated & Fixed\\\bottomrule
\end{tabular}\end{table}

Each generated $\widehat\omega$ is an output of the trained system, rather than an optimizer state fitted to the target. Generation can involve repeated denoiser evaluations without any target-side gradient update. This distinction separates iterative sampling from iterative supervised adaptation.

\section{Adapter and Condition}
\label{sec:app_components}
\subsection{Variable-length block layout}
Within a backbone instantiation, only the active companion set $A\subseteq\mathcal M$
contributes modality-specific parameters:
\begin{equation}\label{eq:app_layout}
\omega=\left[\omega^{\mathrm{sh}};
\{\omega^m_{\mathrm{temp}},\omega^m_{\mathrm{spec}},\omega^m_{\mathrm{proj}}\}_{m\in A}\right]
\in\mathbb R^{P(A)}.
\end{equation}
Each active modality contributes three blocks, with one additional shared block:
\begin{equation}
M=|A|,\qquad B=3M+1,\qquad U\in\mathbb R^{B\times128}.
\end{equation}
Thus $M=1,2,3$ gives $B=4,7,10$, respectively; ten is not a universal block count.
Missing modalities create neither parameter blocks nor placeholder latent tokens.
Block-type and modality conditioning, together with attention masks, support
variable-length sequences while preserving the same functional block schema.

With block size $p_b$, the parameter count is
\begin{equation}
P(A)=p_{\mathrm{sh}}+\sum_{m\in A}
(p^m_{\mathrm{temp}}+p^m_{\mathrm{spec}}+p^m_{\mathrm{proj}})
=\sum_{b=1}^{B}p_b.
\end{equation}
For the active sequence, let $o_b=\sum_{r<b}p_r$ and
$\omega_b=\omega[o_b:o_b+p_b]$.
Offsets are defined within that configuration; semantic block identity is
determined by type and modality, not by a universal vector position.

\subsection{Temporal and spectral branches}
An active companion modality $m\in A$ contributes two feature streams:
\begin{align}
H^m_{\mathrm{temp}}&=T^m(X^m;\omega^m_{\mathrm{temp}}),\label{eq:app_temporal}\\
H^m_{\mathrm{spec}}&=S^m(\mathrm{FFT}(X^m);\omega^m_{\mathrm{spec}}),\\
F^m&=P^m([H^m_{\mathrm{temp}};H^m_{\mathrm{spec}}];\omega^m_{\mathrm{proj}}).
\end{align}
The temporal stream represents variation in the input sequence, while the spectral stream provides frequency-domain information. Projection combines the two streams into the interface consumed by the shared coordinator. The distinct parameter groups allow weight generation to preserve the functional role of each block instead of treating the entire adapter as an unstructured vector.

The companion projection is distinct from the EFM encoder. It changes how additional signals enter the frozen representation, while the original EEG mapping $f$ remains unchanged. The generated weights therefore specify a multimodal extension of an existing representation rather than a replacement EEG backbone.

\subsection{Shared coordination and masking}
The coordinator processes the EEG representation and available companion features to produce an auxiliary residual:
\begin{equation}\label{eq:app_coord}
\Delta Z_{\mathrm{aux}}=K(Z,\{F^m:m\in A\};\omega^{\mathrm{sh}}),\qquad
Z_A=Z+\Delta Z_{\mathrm{aux}}.
\end{equation}
Only available branches contribute. For an empty companion set, the adapter
returns the EEG representation directly:
\begin{equation}\label{eq:app_empty}
a(Z,\varnothing;\omega)=Z,\qquad L_t^{\varnothing}(\omega)=h_t(f(X^{\mathrm{EEG}})).
\end{equation}
For EEG-only inference, the auxiliary residual is zero and the frozen EEG path is unchanged.
Unavailable modalities have no parameter blocks or latent tokens, and therefore
make no contribution to the prediction path.

\subsection{Three condition factors}
The condition encoder uses unlabeled calibration signals and the supported task:
\begin{align}
C_{\mathrm{mod}}&=F_{\mathrm{mod}}(\{V^m+e^m_{\mathrm{mod}}\}_{m\in\{\mathrm{EEG}\}\cup A}),\\
C_{\mathrm{sub}}&=F_{\mathrm{sub}}(q_{\mathrm{sub}},\{V_i^m\}_{m\in\{\mathrm{EEG}\}\cup A,i}),\\
C_\tau&=E_{\mathrm{task}}(\tau;\xi_\tau),\\
C&=E_{\mathrm{cond}}(R^{\mathrm{cond}},\tau;\xi)=F_{\mathrm{cond}}([C_{\mathrm{mod}};C_{\mathrm{sub}};C_\tau]).\label{eq:app_condition}
\end{align}
Here $V_i^m$ is a sample-level feature and $V^m$ aggregates a recording's features for modality $m$. The embeddings $e^m_{\mathrm{mod}}$ identify signal types. The subject query $q_{\mathrm{sub}}$ is task-independent; the task enters separately through $C_\tau$ and the fusion function $F_{\mathrm{cond}}$.

The subject factor is inferred from the observed recording, not retrieved from a target-subject identifier. Its role is to represent recording-specific physiology that is absent from modality availability alone. Shuffling this factor across recordings preserves a collection of feature vectors but breaks their correspondence to the recording for which an adapter is generated, as examined in Appendix~\ref{sec:app_ablations}.

\subsection{Order and identity}
Set pooling follows the permutation-invariant viewpoint of Deep Sets~\citep{deepsets}. Reordering signal-feature/identity pairs leaves the modality factor unchanged. This invariance does not apply to exchanging signal identities while holding the feature values fixed: assigning an ECG identity to an eye-movement feature changes the meaning of the condition.

For attention pooling with a fixed query and consistently permuted key/value pairs, a permutation only reorders the terms in the weighted sum. To see this, write $c=\sum_i a_i v_i$, where $a_i=\exp(s_i)/\sum_j\exp(s_j)$. A permutation $\rho$ gives
\begin{equation}
\sum_i\frac{\exp(s_{\rho(i)})}{\sum_j\exp(s_{\rho(j)})}v_{\rho(i)}
=\sum_i a_i v_i.
\end{equation}
This identity explains the distinction between order invariance and semantic identity. It assumes that scores and values are permuted together, as required for pooling a set of features.

\subsection{Support of the condition}\label{sec:app_condition_support}
The task factor identifies a supported prediction task rather than a dataset identifier. Dataset holdout and task holdout are distinct: a new dataset can share a task with source data, whereas a task-holdout target has no source adapter for its task. FoG is the task-holdout target in our benchmark; its task factor $C_\tau$ is a learned unspecified-task embedding. Similarly, a new combination of supported signals differs from a signal type outside $\mathcal M$. The supported block schema defines the modality interface; it does not by itself create a new encoder for an arbitrary signal type.

\section{Datasets and Additional Results}
\label{sec:app_datasets}
\subsection{Evaluation targets}
The displayed benchmark covers six targets and four task families: ISRUC-S3~\citep{isruc}, HMC~\citep{hmc}, and MASS~\citep{mass} for sleep staging; SEED-V~\citep{seedv} for emotion recognition; EEGMAT~\citep{eegmat} for cognitive-load estimation; and FoG~\citep{fog} for freezing-of-gait detection. This section analyzes the six held-out target datasets in Table~3. The full evaluation additionally includes CAP, DREAMER, and BCI-IV. The CBraMod ablation average in Table~5 and Appendix~\ref{sec:app_ablations} covers all nine target datasets; results on CAP, DREAMER, and BCI-IV are reported in Table~4.

\subsection{Controlled comparisons}
Within each backbone group, the EEG-only model provides the original frozen prediction path. Mean source weights and nearest source weights reuse fitted source adapters; Direct MLP predicts weights from the condition. These methods and ZeroMAG use the same frozen EEG encoder and head. The supervised reference is a separate multimodal fine-tuning experiment using the labeled target training split and all available modalities. Its purpose is to measure the gap between ZeroMAG and supervised adaptation; it is not a step in ZeroMAG training or inference and is not a mathematical upper bound.

The cross-paradigm comparisons use PhysioOmni~\citep{physioomni}, the original ICML 2024 SleepFM~\citep{sleepfm2024}, and Brant-X~\citep{brantx}. These models have native backbones and different multimodal learning procedures. Their scores are the evaluations reported in this paper, rather than scores taken from the cited publications. In particular, the SleepFM reference here is the original \texttt{sleepfm-codebase} model, not the later clinical model.

\subsection{Results by task family}
The task-family comparisons below use Table~3; the same numerical cells are not repeated here.

\paragraph{Sleep staging}
ZeroMAG improves over EEG-only by 7.48--7.87 points on ISRUC-S3 and HMC, and by 7.74--8.01 points on MASS. Its smallest supervised-reference gap in the displayed benchmark is 0.38 points for CBraMod on MASS. Improvements occur for all three separately trained backbone instantiations, although their absolute scores differ.

\paragraph{Emotion recognition}
SEED-V has smaller absolute improvements over EEG-only, at 4.21, 4.36, and 4.32 points for CBraMod, CSBrain, and CodeBrain. Its recovery of supervised gain is also lower than that of the other displayed targets, ranging from 87.34\% to 88.71\%. Thus, the lower gain is not explained solely by a smaller supervised-minus-EEG baseline interval. The table does not identify the source of this task-dependent gap.

\paragraph{Cognitive load and gait}
On EEGMAT, the improvements over EEG-only are 7.61, 7.70, and 7.89 points. The corresponding FoG improvements are 7.86, 8.64, and 7.88 points. CSBrain on FoG has the largest displayed gain over EEG-only, whereas CodeBrain has the highest absolute ZeroMAG score on both targets. The best absolute score and the largest incremental gain therefore need not identify the same backbone.

\subsection{Gains across backbones}
\begin{figure}[htbp]\centering
\includegraphics[width=\linewidth]{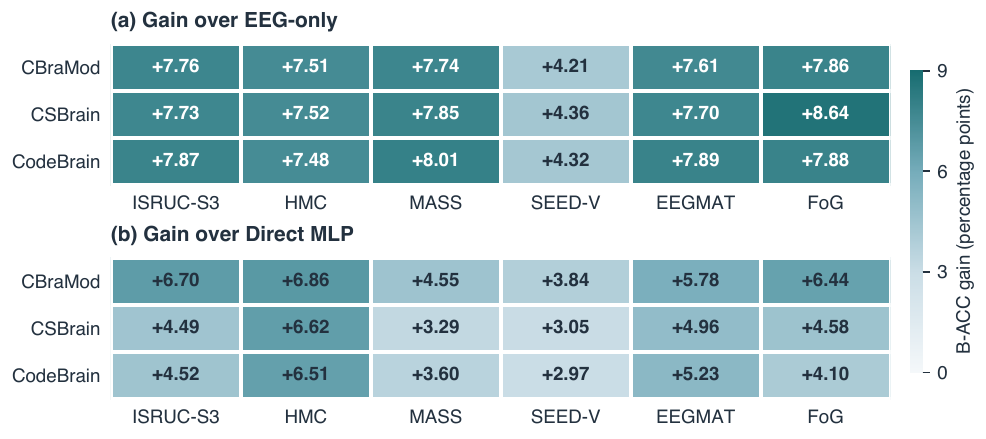}
\caption{Gains in balanced accuracy over EEG-only and Direct MLP, in percentage points. Both panels use the same scale. Values are differences of means from Table~3; no paired uncertainty or significance is inferred.}\label{fig:app_gains}
\end{figure}

Fig.~\ref{fig:app_gains} compares the improvement over EEG-only and Direct MLP with a common color scale. All 18 backbone--dataset settings have positive mean differences for both comparisons. The largest advantage over Direct MLP occurs on HMC with CBraMod (6.86 points); the smallest occurs on SEED-V with CodeBrain (2.97 points).

\subsection{Aggregate comparisons}

Let $e_j,z_j,s_j,m_j$ be EEG-only, ZeroMAG, supervised, and Direct MLP balanced-accuracy means for setting $j$. Values are in percentage points. Define
\begin{equation}
 \Delta^{\mathrm{EEG}}_j=z_j-e_j,\qquad
 \Delta^{\mathrm{MLP}}_j=z_j-m_j,\qquad
 G^{\mathrm{sup}}_j=s_j-z_j.
\end{equation}
Table~\ref{tab:derived_gains} lists differences between reported means. Paired uncertainty requires the seed-wise covariance.

\begin{table}[htbp]\centering\small
\caption{Arithmetic derived from Table~3; no additional experimental runs. Recovery is $100(z-e)/(s-e)$ for each row.}\label{tab:derived_gains}
\begin{tabular}{llrrrr}
\toprule
Backbone & Target & $\Delta^{\mathrm{EEG}}$ & $\Delta^{\mathrm{MLP}}$ & $G^{\mathrm{sup}}$ & Recovery (\%)\\
\midrule
CBraMod & ISRUC-S3 & 7.76 & 6.70 & 0.45 & 94.52 \\
CBraMod & HMC & 7.51 & 6.86 & 0.52 & 93.52 \\
CBraMod & MASS & 7.74 & 4.55 & 0.38 & 95.32 \\
CBraMod & SEED-V & 4.21 & 3.84 & 0.61 & 87.34 \\
CBraMod & EEGMAT & 7.61 & 5.78 & 0.47 & 94.18 \\
CBraMod & FoG & 7.86 & 6.44 & 0.55 & 93.46 \\
CSBrain & ISRUC-S3 & 7.73 & 4.49 & 0.43 & 94.73 \\
CSBrain & HMC & 7.52 & 6.62 & 0.56 & 93.07 \\
CSBrain & MASS & 7.85 & 3.29 & 0.41 & 95.04 \\
CSBrain & SEED-V & 4.36 & 3.05 & 0.58 & 88.26 \\
CSBrain & EEGMAT & 7.70 & 4.96 & 0.49 & 94.02 \\
CSBrain & FoG & 8.64 & 4.58 & 0.53 & 94.22 \\
CodeBrain & ISRUC-S3 & 7.87 & 4.52 & 0.48 & 94.25 \\
CodeBrain & HMC & 7.48 & 6.51 & 0.50 & 93.73 \\
CodeBrain & MASS & 8.01 & 3.60 & 0.42 & 95.02 \\
CodeBrain & SEED-V & 4.32 & 2.97 & 0.55 & 88.71 \\
CodeBrain & EEGMAT & 7.89 & 5.23 & 0.51 & 93.93 \\
CodeBrain & FoG & 7.88 & 4.10 & 0.60 & 92.92 \\
\bottomrule
\end{tabular}
\end{table}

\subsubsection{Aggregate gain recovery}
Averaging the 18 displayed means with equal weight yields 69.6133 for EEG-only, 76.8322 for ZeroMAG, 77.3344 for the supervised reference, and 71.9383 for Direct MLP. The corresponding gains are 7.22 and 4.89 points, with a 0.50-point supervised gap. The corresponding advantages over mean and nearest source weights are 6.7772 and 5.9989 points, rounding to 6.78 and 6.00.

Recovery can be aggregated in two ways:
\begin{align}
 R_{\mathrm{ratio}}&=100\frac{\sum_j(z_j-e_j)}{\sum_j(s_j-e_j)},\\
 R_{\mathrm{mean}}&=\frac{100}{J}\sum_j\frac{z_j-e_j}{s_j-e_j}.
\end{align}
The ratio of aggregate gains is 93.50\%; averaging row-wise recovery gives 93.12\%. We use the ratio of aggregate gains for the 93.50\% summary. The row-wise ratios are listed separately in Table~\ref{tab:derived_gains}. All displayed supervised gains are positive.

The fraction of the Direct MLP gap closed by ZeroMAG is
\begin{equation}
 100\frac{\sum_j(z_j-m_j)}{\sum_j(s_j-m_j)}=90.6929\%,
\end{equation}
This 90.69\% value measures the fraction of the supervised-minus-MLP performance gap closed by ZeroMAG.

\subsubsection{Backbone comparison}
The mean improvement over EEG-only is 7.1150 points for CBraMod, 7.3000 for CSBrain, and 7.2417 for CodeBrain. Their rounded values are 7.12, 7.30, and 7.24. Their range is 0.185 points (approximately 0.19 points), calculated before rounding the backbone means.

The backbone-wise ratios of aggregate gains are 93.47\%, 93.59\%, and 93.42\%. Across the 18 rows, the supervised-minus-ZeroMAG margin ranges from 0.38 to 0.61 points. ZeroMAG's displayed mean is higher than the EEG-only, source-transfer, and Direct MLP means in every setting. Significance is not inferred from these means.

\subsection{Reported variation and the supervised gap}
\begin{figure}[htbp]\centering
\includegraphics[width=\linewidth]{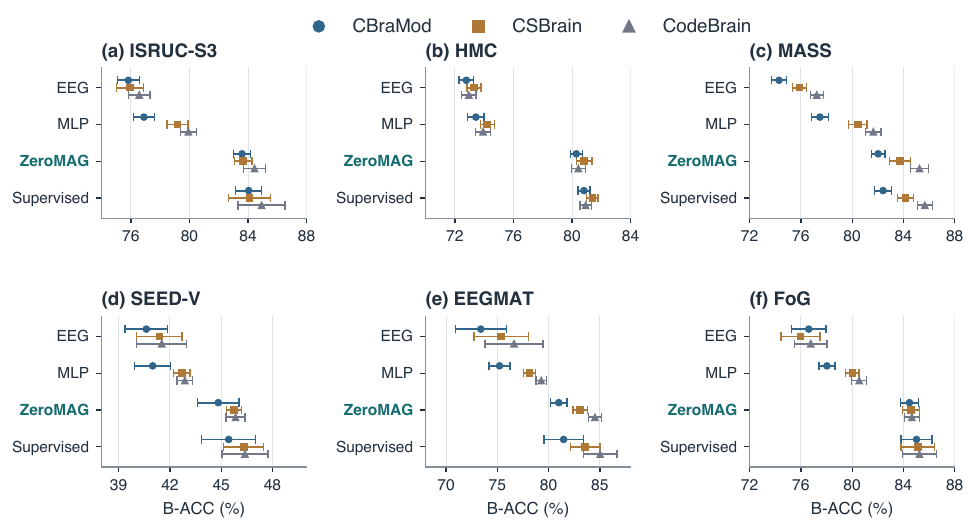}
\caption{Means and reported $\pm1$ SD for four methods within each target. Marker shapes identify backbones. Horizontal limits vary by target to show within-target differences; all values are balanced accuracy (\%).}\label{fig:app_intervals}
\end{figure}

Fig.~\ref{fig:app_intervals} preserves the reported standard deviations for EEG-only, Direct MLP, ZeroMAG, and supervised fitting. These summarize seed-to-seed variation. They are not confidence intervals for pairwise differences, and overlapping whiskers do not establish either significance or equivalence.

The supervised-reference gap ranges from 0.38 to 0.61 points across the 18 settings, with mean 0.5022 points. This is a small residual difference in the displayed averages, but it is not a formal equivalence result. The corresponding gain-recovery ratios describe the position of ZeroMAG between EEG-only and supervised fitting, rather than the probability of recovering the fitted adapter parameters.

\subsection{Cross-paradigm comparisons}
With CodeBrain, ZeroMAG exceeds the original SleepFM on all six targets by an average of 1.8533 points. The average gaps to PhysioOmni and Brant-X are 0.3050 and 0.8033 points, respectively. These comparisons place the generated-adapter framework alongside native multimodal models; they do not isolate backbone architecture, pretraining exposure, or target adaptation as a single controlled factor.

PhysioOmni explicitly supports missing modalities, while Brant-X studies physiological-signal alignment and SleepFM learns multimodal sleep representations. The distinguishing question here is how a separately trained generator equips an existing frozen EEG model with an adapter. Missing-modality support alone does not separate these approaches.

\section{Training and Deployment}
\label{sec:app_procedure}
\subsection{Two training stages}
ZeroMAG has two learned stages followed by deployment. Stage 1 learns the adapter representation and decoder; Stage 2 learns the conditional diffusion model and condition encoder while retaining the Stage 1 representation. Fig.~3 shows their inputs, outputs, and update boundaries. The target deployment path is an inference procedure, not a third training stage.

\subsection{Reference adapters}
For a source recording, the reference adapter solves the supervised fitting problem with the EFM encoder and head fixed:
\begin{equation}\label{eq:app_reference}
\omega^*=\arg\min_{\omega\in\mathbb R^{P(A)}}\sum_{r_i\in R^{\mathrm{fit}}}\ell(L_i^A(\omega),Y_i).
\end{equation}
Its weights provide the parameter target for representation learning and conditional generation. Its predictions on the disjoint source subset $R^{\mathrm{fun}}$ provide the functional target. Thus, reference fitting and functional supervision use different source samples, and the latter still belongs to offline training rather than to target evaluation.

\subsection{Stage 1: representation learning}
For each block, the encoder predicts a Gaussian mean and variance, samples a pre-contextual token, and applies the cross-block transformer:
\begin{align}
U_{b,\mathrm{pre}}&=\mu_b+\exp(\ell_b/2)\odot\epsilon_b,
\qquad\epsilon_b\sim\mathcal N(0,I),\\
U&=\mathrm{CBT}([U_{b,\mathrm{pre}}+E_b]_{b=1}^B),\\
\widetilde\omega&=d(U;\psi).
\end{align}
The representation follows variational encoding~\citep{kingma2013auto}; the shared block schema and functional objective specify how it is applied to adapter weights. Stage 1 combines block reconstruction, direction alignment, functional agreement, posterior regularization, and expert balancing~\citep{fedus2022switch}:
\begin{equation}\label{eq:app_ae}
\mathcal L_{\mathrm{AE}}=\mathcal L_{\mathrm{blk}}+0.1\mathcal L_{\mathrm{cos}}
+10^{-4}\mathcal L_{\mathrm{KL}}+1.0\delta_{\mathrm{fun}}(\omega^*,\widetilde\omega)
+0.01\mathcal L_{\mathrm{bal}}.
\end{equation}
The reconstruction term acts on block parameters, the cosine term on direction, the KL term on the Gaussian posterior, and the balancing term on expert use. Functional supervision constrains the decoded adapter's predictions through the frozen EFM. The terms address different properties and do not turn the whole objective into a pure likelihood bound.

\begin{algorithm}[htbp]
\caption{Stage 1: adapter representation learning}\label{alg:app_stage1}
\begin{algorithmic}[1]
\Require Source references $\omega^*$ and associated $R^{\mathrm{fun}}$; fixed $f,h$.
\Ensure Learned encoder $\eta$, decoder $\psi$, and canonical targets $U_0$.
\While{Stage 1 training continues}
\State Sample source reference adapters and associated functional inputs.
\State Encode blocks into $(\mu_b,\ell_b)$ and sample $U_{b,\mathrm{pre}}$.
\State $U\gets\mathrm{CBT}([U_{b,\mathrm{pre}}+E_b]_{b=1}^B)$.
\State $\widetilde\omega\gets d(U;\psi)$.
\State Evaluate $\mathcal L_{\mathrm{AE}}$ in Eq.~\eqref{eq:app_ae}.
\State Update $\eta,\psi$; keep $f,h,\omega^*$ fixed.
\EndWhile
\State Freeze $\eta,\psi$.
\For{each source reference $\omega^*$}
\State $U_0\gets\bar e(\omega^*;\eta)=\mathrm{CBT}([\mu_b+E_b]_{b=1}^B)$.
\EndFor
\end{algorithmic}\end{algorithm}

\subsection{Stage 2: conditional generation}
The canonical target fixes a deterministic coordinate for each reference adapter:
\begin{equation}\label{eq:app_canonical}
U_0=\bar e(\omega^*;\eta)=\mathrm{CBT}([\mu_b+E_b]_{b=1}^B).
\end{equation}
The condition $C=E_{\mathrm{cond}}(R^{\mathrm{cond}},\tau;\xi)$ guides the diffusion-transformer noise predictor~\citep{peebles2023scalable}. The noise-prediction, latent, weight, and functional terms jointly train $\theta$ and $\xi$. The latter three are active only when $\bar\alpha_t>0.5$:
\begin{equation}\label{eq:app_gen_configured}
\mathcal L_{\mathrm{gen}}=\mathcal L_{\mathrm{diff}}
+\mathbf 1[\bar\alpha_t>0.5]\left(
0.1\mathcal L_{\mathrm{lat}}+0.1\mathcal L_{\mathrm w}
+1.0\delta_{\mathrm{fun}}(\omega^*,\widehat\omega)\right).
\end{equation}
The diffusion term is active at every timestep. Expert balancing is used only
in Stage~1; neither stage uses a diversity loss.

\begin{algorithm}[htbp]
\caption{Stage 2: conditional adapter generation}\label{alg:app_stage2}
\begin{algorithmic}[1]
\Require Source triples $(R,\tau,\omega^*)$; fixed $\eta,\psi,f,h$ and $U_0$.
\Ensure Learned denoiser $\theta$ and condition encoder $\xi$.
\While{Stage 2 training continues}
\State Sample a source triple and unlabeled $R^{\mathrm{cond}}\subseteq R^{\mathrm{fit}}$.
\State $C\gets E_{\mathrm{cond}}(R^{\mathrm{cond}},\tau;\xi)$.
\State Sample a training timestep $t$ and $\epsilon\sim\mathcal N(0,I)$.
\State $U_t\gets\sqrt{\bar\alpha_t}U_0+\sqrt{1-\bar\alpha_t}\epsilon$.
\State Predict $\epsilon_\theta(U_t,t,C)$ and evaluate $\mathcal L_{\mathrm{diff}}$.
\If{$\bar\alpha_t>0.5$}
\State Estimate $\widehat U_0$ using Eq.~\eqref{eq:app_clean}.
\State $\widehat\omega\gets d(\widehat U_0;\psi)$.
\State Add weighted latent, weight, and functional losses.
\EndIf
\State Update $\theta,\xi$, propagating auxiliary gradients through fixed $d$.
\EndWhile
\end{algorithmic}\end{algorithm}

\subsection{Deployment}
For an unseen recording, generation requires only the unlabeled calibration subset, the supported task, and the selected backbone's trained generator and decoder. No reference adapter is required. Query samples are excluded from condition construction.

\begin{algorithm}[htbp]
\caption{Target inference with fixed parameters}\label{alg:app_inference}
\begin{algorithmic}[1]
\Require Unlabeled $R^{\mathrm{cond}}$, task $\tau$, query inputs $R^{\mathrm{qry}}$; fixed model components.
\Ensure Query predictions with a generated adapter.
\State $C\gets E_{\mathrm{cond}}(R^{\mathrm{cond}},\tau;\xi)$.
\State Draw $\epsilon\sim\mathcal N(0,I)$.
\State $\widehat\omega\gets d(s_\theta(\epsilon,C);\psi)$.
\For{each query sample $(X_i^{\mathrm{EEG}},X_i^A)$}
\State $Z_i\gets f(X_i^{\mathrm{EEG}})$.
\State $\widehat Y_i\gets\arg\max h(a(Z_i,X_i^A;\widehat\omega))$.
\EndFor
\end{algorithmic}\end{algorithm}

The sampled initial noise makes adapter generation stochastic even when the remaining inference operations are fixed. This differs from posterior sampling during Stage 1: deployment samples a conditional adapter latent, whereas Stage 1 samples a representation of a known reference vector. Neither process requires target labels.

\section{Objective Analysis}
\label{sec:app_derivations}
\subsection{Canonical latent coordinates}
The block posterior is defined before cross-block contextualization. Let $F_\eta$ denote adding block embeddings and applying CBT. For pre-contextual tokens $U_{\mathrm{pre}}$, the stochastic output is $F_\eta(U_{\mathrm{pre}})$ and the canonical output is $F_\eta(\mu)$. In general,
\begin{equation}
F_\eta(\mathbb E[U_{\mathrm{pre}}])\ne\mathbb E[F_\eta(U_{\mathrm{pre}})].
\end{equation}
For example, a scalar nonlinear map $F(u)=u^2$ gives $F(\mathbb E[u])=\mu^2$ but $\mathbb E[F(u)]=\mu^2+\sigma^2$. The same distinction applies to nonlinear contextualization. The Stage 2 target is the explicit means-before-CBT path in Eq.~\eqref{eq:app_canonical}, not an estimate of the exact mean after a nonlinear transformer.

\subsection{Posterior regularization}
For a diagonal Gaussian posterior $q_b=\mathcal N(\mu_b,\operatorname{diag}(\exp\ell_b))$, its KL to a standard normal has the closed form
\begin{equation}
D_{\mathrm{KL}}(q_b\Vert\mathcal N(0,I))=
\frac12\sum_{r=1}^{d}(\mu_{b,r}^2+\exp\ell_{b,r}-1-\ell_{b,r}).
\end{equation}
Here $d$ is token width and $r$ indexes a token coordinate. This identity concerns the pre-contextual posterior. It does not imply that the distribution after CBT is diagonal Gaussian. It also does not specify whether a training implementation sums or averages over blocks; a change of reduction changes the scale of the corresponding loss coefficient.

For one coordinate, the partial derivatives are $\partial D_{\mathrm{KL}}/\partial\mu=\mu$ and $\partial D_{\mathrm{KL}}/\partial\ell=(\exp\ell-1)/2$. The minimum occurs at zero mean and unit variance when this term is considered alone. Reconstruction and functional objectives compete with that regularizer, so a trained posterior is not expected to minimize the KL term independently of the other losses.

\subsection{Block reconstruction and direction}
Parameter reconstruction and cosine alignment provide distinct constraints. A positive rescaling of a nonzero reference block, $\widetilde\omega_b=c\omega_b^*$ with $c>0$, has cosine similarity one even when $c\ne1$. Its squared reconstruction error is $(c-1)^2\|\omega_b^*\|_2^2$. Conversely, a small error in a large block can dominate a coordinate-wise reconstruction sum while a smaller block has a larger relative discrepancy.

Thus, block structure carries information that a whole-vector direction score alone does not recover. The reconstruction and direction objectives complement predictive supervision through the frozen EFM. Neither imposes a direct bound on target accuracy.

\subsection{Functional discrepancy and its derivative}
On a fixed source functional subset, write the reference probability vector as $p_i=\operatorname{softmax}(L_i^A(\omega^*))$ and the student vector as $q_i=\operatorname{softmax}(L_i^A(\omega'))$. The main discrepancy is
\begin{equation}
 \delta_{\mathrm{fun}}(\omega^*,\omega')=
 \frac{1}{n}\sum_{i=1}^{n}\sum_{c=1}^{K}
 p_{i,c}\log\frac{p_{i,c}}{q_{i,c}}.
\end{equation}
Reference probabilities are constants for the student update. For a student logit $z_{i,c}$, the derivative is
\begin{equation}
 \frac{\partial\delta_{\mathrm{fun}}}{\partial z_{i,c}}
 =\frac{1}{n}(q_{i,c}-p_{i,c}).
\end{equation}
This follows by expressing the cross-entropy term as $-\sum_cp_cz_c+\log\sum_c\exp z_c$ and using $\sum_cp_c=1$. It shows that the objective aligns distributions, not just the most likely class.

No labels are required to compute this discrepancy. Nonetheless, its reference predictions inherit the behavior of an adapter that was fitted using source labels. Functional matching is therefore not an unsupervised proof of correctness: it transfers the source reference's useful behavior and its errors. A confident but wrong teacher prediction is still a target of the KL term.

The direction of KL also matters. The main text uses reference-to-student KL; reversing its arguments produces a different objective. The expression uses the untempered softmax in the main formulation. Stable log-softmax evaluation preserves this objective without taking logarithms of rounded probabilities.

\subsection{Prediction agreement}
A small parameter distance alone does not ensure a small predictive error. Locally, for a differentiable logit map $L$,
\begin{equation}
 L(\omega+\Delta\omega)-L(\omega)
 =J_L(\omega)\Delta\omega+o(\norm{\Delta\omega}_2).
\end{equation}
Sensitivity depends on the Jacobian and on the direction of the error. Two equally sized perturbations can have different functional effects. Conversely, multiple parameter vectors can implement the same predictions, so exact weight recovery is not necessary for functional agreement.

Prediction agreement admits a simple margin condition. Suppose $p$ and $q$ are two class-probability vectors, the reference winning class is $c^*$, and its margin is
\begin{equation}
 m=p_{c^*}-\max_{c\ne c^*}p_c>0.
\end{equation}
If $\norm{p-q}_\infty<m/2$, the student has the same unique winning class. Indeed, for any competitor $c$,
\begin{equation}
 q_{c^*}-q_c\ge p_{c^*}-p_c-2\norm{p-q}_\infty>0.
\end{equation}
This statement concerns agreement with the reference, not agreement with the true label. It also does not convert a mean training-subset KL into a uniform target-query bound. A new dataset can shift both margins and discrepancies. Functional consistency is an objective evaluated through predictive behavior, rather than a universal guarantee of target generalization.

\subsection{Forward diffusion and clean-latent estimation}
Following the Gaussian diffusion formulation~\citep{ddpm}, for schedule coefficients $\alpha_t=1-\beta_t$ and $\bar\alpha_t=\prod_{r=1}^{t}\alpha_r$, the forward marginal is
\begin{equation}
 U_t=\sqrt{\bar\alpha_t}U_0+\sqrt{1-\bar\alpha_t}\epsilon,
 \qquad \epsilon\sim\mathcal N(0,I).
\end{equation}
The conditional denoiser predicts $\epsilon_\theta(U_t,t,C)$. Solving the forward relation for $U_0$ after substituting this prediction gives
\begin{equation}
 \label{eq:app_clean}\widehat U_0=
 \frac{U_t-\sqrt{1-\bar\alpha_t}\epsilon_\theta(U_t,t,C)}
 {\sqrt{\bar\alpha_t}}.
\end{equation}
No new assumption is needed for this algebra beyond $\bar\alpha_t>0$. The estimator's quality does depend on learned noise prediction and numerical stability.

\subsection{Low-noise supervision}
Subtract the true $U_0$ from its estimator to obtain the exact identity
\begin{equation}
 \widehat U_0-U_0=
 \sqrt{\frac{1-\bar\alpha_t}{\bar\alpha_t}}
 \bigl(\epsilon-\epsilon_\theta(U_t,t,C)\bigr).
\end{equation}
Thus squared clean-latent error multiplies squared noise-prediction error by $(1-\bar\alpha_t)/\bar\alpha_t$. This factor increases as the signal coefficient decreases. Low-noise gating in the main text is therefore consistent with avoiding strongly amplified reconstruction and functional gradients from heavily corrupted latents.

Writing $\mathrm{SNR}_t=\bar\alpha_t/(1-\bar\alpha_t)$, the multiplier is $1/\mathrm{SNR}_t$. For a fixed noise-prediction residual, a lower signal-to-noise ratio therefore yields a larger clean-latent error. This algebra motivates applying the auxiliary losses at low-noise timesteps; the implementation uses $\bar\alpha_t>0.5$ (SNR $>1$) with a cosine noise schedule, as specified in Appendix~\ref{sec:app_implementation}.

\subsection{MoE decoding}
Sparse mixture-of-experts models select a subset of experts for each input~\citep{sparsemoe}. In ZeroMAG, routing uses the contextualized token and its block identity:
\begin{align}
\pi_b&=\mathrm{softmax}(\mathrm{route}([U_b;E_b];\psi_r)),\\
\widetilde\omega_b&=\mathrm{head}_{\mathrm{type}(b)}\left(\sum_{j\in\mathrm{TopK}(\pi_b)}\pi_{b,j}r_j(U_b)\right).
\end{align}
The block type is temporal, spectral, projection, or shared. Experts act on latent tokens; the output heads map their mixtures back to the appropriate parameter blocks. The mixture uses the softmax routing weights $\pi_{b,j}$ for the selected experts, as in the main formulation.

Top-$k$ selection restricts the active expert set. Away from routing-boundary changes, the selected computation is differentiable through expert outputs and routing weights. At a boundary, the selected set can change discontinuously. The presence of an expert-balancing objective does not establish uniform expert utilization; the main formulation uses it as a training regularizer.

\subsection{Gradients through a frozen decoder and EFM}
Write the student logit path as $z=h(a(f(X),X^A;d_\psi(\widehat U_0)))$. For an auxiliary loss $\mathcal J(z)$, the chain rule includes
\begin{equation}
 \frac{\partial\mathcal J}{\partial\theta}
 =\frac{\partial\mathcal J}{\partial z}
 \frac{\partial z}{\partial\widehat\omega}
 \frac{\partial d_\psi(\widehat U_0)}{\partial\widehat U_0}
 \frac{\partial\widehat U_0}{\partial\theta}.
\end{equation}
Freezing $\psi$ means not updating its stored values. It does not set its input Jacobian to zero. Similarly, the head can remain frozen while its derivative transmits a functional error to generated adapter weights. Reference predictions may be detached; generated student predictions must preserve the relevant graph.

More explicitly, the Stage 2 parameter update satisfies $\psi^{(k+1)}=\psi^{(k)}$, whereas $\theta$ and $\xi$ receive gradients. This fixed-parameter condition does not require the mathematical derivative $\partial\mathcal J/\partial\psi$ to vanish. The relevant derivative for generator learning is the decoder input Jacobian, which remains in the chain rule.

For the condition encoder, the same path additionally contains $\partial\widehat U_0/\partial C$ and $\partial E_{\mathrm{cond}}/\partial\xi$. Thus the functional objective can shape the recording-derived condition even though the teacher, decoder, and EFM are fixed. Detaching the generated weights would eliminate this auxiliary supervision rather than merely freezing the decoder.

\subsection{Latent error and functional error}
Let $\widehat U_0=U_0+\Delta U$ and assume the decoder is differentiable at $U_0$. Locally,
\begin{equation}
d(U_0+\Delta U;\psi)-d(U_0;\psi)=J_d(U_0)\Delta U+o(\|\Delta U\|).
\end{equation}
Combining this with the logit Jacobian gives a first-order predictive perturbation $J_LJ_d\Delta U$. Its size depends on both decoder sensitivity and the frozen EFM's response to the decoded adapter. Equal latent reconstruction errors can therefore yield different logit errors, motivating an objective that observes the final prediction path.

There are also two distinct discrepancies: reconstruction compares $d(\bar e(\omega^*))$ with $\omega^*$, while generation compares $d(\widehat U_0)$ with $\omega^*$. A generator that exactly recovers $U_0$ still inherits any reconstruction error in the learned representation. Conversely, a high-quality autoencoder does not ensure that the conditional diffusion model recovers the appropriate latent for every recording.

\section{Efficiency}
\label{sec:app_efficiency}
\subsection{Measured target-side time}
Target-side wall-clock time was measured programmatically with timers in the code, under the same hardware and protocol for the two compared procedures. Condition construction and adapter generation took approximately 2 minutes, compared with approximately 15 minutes for supervised target adaptation. These measurements give
\begin{equation}
\mathrm{speedup}=\frac{15}{2}=7.5,\qquad
\mathrm{time\ reduction}=100\left(1-\frac{2}{15}\right)\approx86.7\%.
\end{equation}
The difference is approximately 13 minutes per measured adaptation procedure. These are elapsed-time measurements of the reported procedures, rather than isolated GPU kernel timings. The timed procedures exclude source-bank construction and offline generator training.

\subsection{Cost decomposition}
The target workflow consists of conditioning, iterative denoising, decoding, and installing the generated weights. A symbolic decomposition is
\begin{equation}
T_{\mathrm{gen}}=T_{\mathrm{cond}}+N_{\mathrm{eval}}T_{\mathrm{denoise}}
+T_{\mathrm{decode}}+T_{\mathrm{install}}.
\end{equation}
Here $N_{\mathrm{eval}}$ is the number of denoiser evaluations and each $T$ denotes the corresponding elapsed cost under a fixed execution convention. This expression describes the operations; it does not assign portions of the measured time to individual components. Query prediction then adds the cost of the fixed EEG model and active adapter branches.

No target-side optimization is required, but generation still performs computation. The 7.5$\times$ measurement compares adaptation procedures; it does not establish a 7.5$\times$ speedup for every subsequent query prediction or for the entire offline-plus-online pipeline.

\subsection{Amortization across targets}
Let $T_{\mathrm{off}}$ denote source-bank construction and both training stages for one separately trained backbone instantiation. For $N$ deployments, the generated-adapter route has cost $T_{\mathrm{off}}+NT_{\mathrm{gen}}$, excluding shared query prediction. The supervised route has incremental adaptation cost $NT_{\mathrm{sup}}$. Under equal per-target costs, the generated route becomes cheaper when
\begin{equation}
N>\frac{T_{\mathrm{off}}}{T_{\mathrm{sup}}-T_{\mathrm{gen}}}.
\end{equation}
Using the measured target-side difference gives $N>T_{\mathrm{off}}/(13\ \mathrm{minutes})$. This is a symbolic break-even condition, not a measured deployment count: no numerical offline training cost is assigned here. Offline training is associated with a fixed source configuration and backbone; separate backbone instantiations have separate offline costs.

\subsection{Latent and output complexity}
For $B$ tokens of width $d$, dense latent self-attention stores an attention map with $B^2$ entries per head. Token projections and feed-forward transformations also depend on $B$ and $d$. Reducing latent width can reduce the denoiser computation, while the decoder still produces $P$ scalar adapter coordinates. Compact latent generation does not remove the output storage and installation cost of the adapter itself.

Sparse expert selection limits the active reconstruction functions in the stated decoder. The runtime benefit depends on how expert computation is executed; the selection rule alone does not provide a measured speedup. Functional supervision adds an adapter-and-head prediction path during offline training, but this loss evaluation is absent from target deployment.

\section{Component Ablations}
\label{sec:app_ablations}
\label{sec:app_generation_ablation}
\label{sec:appendix_adapter_heterogeneity}
\subsection{Component ablations}
\label{sec:app_batch_scope}
The main performance comparison and the component ablations were conducted in independent experimental batches for different purposes. The CBraMod nine-target mean from Tables~3 and~4 is 70.01\%, whereas Full ZeroMAG in the ablation batch is $73.41\pm0.71$\% (Table~5). These absolute scores are not a paired cross-batch comparison. All ablation decreases use Full ZeroMAG from the same batch as their reference.

The eight component-ablation drops are 11.09 for task+modality only, 9.24 for shuffled subject condition, 11.07 for direct weight MLP, 8.18 for modular latent MLP, 5.07 for modular latent DiT without MoE, 4.16 for a shared modality encoder, 8.12 for temporal-only encoding, and 6.58 for mean fusion. Functional supervision is evaluated through controlled comparisons within the ablation batch, as detailed in Section~5.6.

The reported task+modality reduction of 11.09 points corresponds to $100\times11.09/73.41=15.1069\%$ of the full score. This is a relative score reduction, not a 15.1 percentage-point change and not a loss of 15.1\% of the supervised gain.

\paragraph{Ablation settings} The source bank, source training splits, frozen EEG model, prediction head, and evaluation protocol are held fixed as stated in the main text. The condition variants test whether recording-specific information and its alignment are useful. The generator variants compare direct parameter regression, latent regression, diffusion, and MoE decoding. The adapter variants alter modality-specific encoding, spectral information, and shared coordination.

\paragraph{Results analysis} The modular latent MLP exceeds direct weight MLP by 2.89 points; replacing that latent regressor with DiT contributes another 3.11 points in the displayed sequence, and the full configuration exceeds the no-MoE variant by 5.07 points. These are incremental differences between the listed variants, not an additive causal decomposition independent of training configuration. Removing subject conditioning and shuffling it cause substantial declines, consistent with recording-aligned information contributing beyond task and modality identity.

\begin{figure}[!t]
\centering
\includegraphics[width=\linewidth]{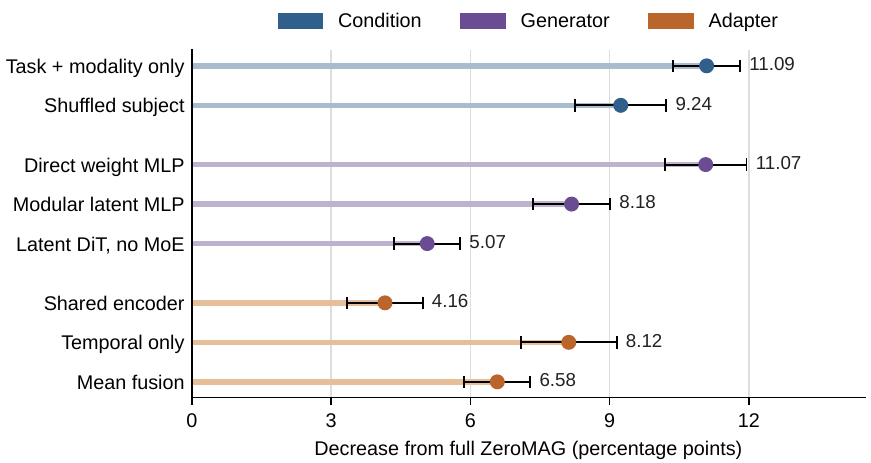}
\caption{\textbf{Performance decrease under component ablation.} Markers show the decreases reported in Table~5 relative to full ZeroMAG (B-ACC $73.41\pm0.71$). Whiskers reproduce the table's reported standard deviations of the decreases; they are not newly calculated confidence intervals. Variants are grouped by condition, generator, and adapter. These are separate interventions, not additive contributions. The aggregation covers all nine target datasets, as in Table~5.}
\label{fig:app_ablation}
\end{figure}

\subsection{Interpreting the interventions}
The shuffled-subject comparison changes alignment between a condition and its recording while retaining the set of subject-condition vectors. Its 9.24-point decrease is consistent with the importance of recording-aligned information. The task-plus-modality variant removes that factor and decreases accuracy by 11.09 points. The two interventions answer related but distinct questions: availability of subject information and correctness of its assignment.

The temporal-only variant has an 8.12-point decrease, compared with 4.16 points for a shared modality encoder and 6.58 points for mean fusion. The larger temporal-only decrease is consistent with a useful contribution from the spectral branch in this experiment. It does not quantify an independent spectral contribution that can be added to the other decreases, because the variants are separate trained systems.

The reported decreases are paired with the standard deviations given in Table~5. They are computed from seed-wise paired differences rather than from the two marginal score standard deviations. The functional-loss ablation in the main article isolates the supervision applied in the two learning stages.

\subsection{Functional supervision}
The functional-supervision variants use five matched random seeds within the same ablation batch, with the controls specified in Section~5.6. Removing both functional-supervision terms reduces B-ACC from $73.41\pm0.71$\% to $70.54\pm0.86$\%, a decrease of 2.87 percentage points. This within-batch comparison supports the contribution of functional supervision; it does not rely on the absolute difference between the independent main-comparison and ablation batches. Normalized-error definitions and results are in Table~6.

\section{Implementation and Evaluation Protocol}
\label{sec:app_protocol}
\subsection{Implementation Details}
\label{sec:app_implementation}
Table~2 gives the training and inference settings.
Batch size counts reference-adapter samples per optimizer step, not subjects
or signal windows. Each sample has $B=3M+1$ active blocks and a $B\times128$
latent sequence. Block identity conditioning and attention masks handle
configuration-dependent sequence lengths without allocating missing-modality blocks.

\subsection{Separate backbone instantiations}
Each of CBraMod, CSBrain, and CodeBrain has a separately trained generator. The EFM encoder is loaded from its checkpoint, and the compatible prediction head is fixed throughout the ZeroMAG pipeline (Section~3.3). Reference-adapter fitting updates only adapter parameters. Functional losses use the same frozen head, whose parameters are not part of the adapter bank or generator output. At deployment, the known task selects its output layer; no target labels or target-side updates are used.

Within a selected instantiation, generated vectors use the block layout expected by its decoder and adapter. This scope is essential for the meaning of $\widehat\omega=g(R,\tau)$: the symbol $g$ identifies that instantiation's generator, rather than one universal weight generator shared by all architectures. The reported multi-backbone evaluation tests repeated applicability of the framework.

\paragraph{FoG prediction-head reuse}\label{sec:app_fog_head}
The FoG experiment uses the existing prediction head from the original method's official implementation without retraining it. ACC is already part of the supported modality set. ZeroMAG's modality encoders and generated adapter extract and adapt auxiliary-signal features, and the resulting representation is passed to the frozen head using the input interface defined by that implementation. This experiment evaluates auxiliary-modality adaptation rather than training a new FoG classifier. As in the other evaluations, the prediction head remains frozen while ZeroMAG adapts the auxiliary-signal representation.

\subsection{Dataset and recording splits}\label{sec:app_splits}
The dataset lists and disjoint source/calibration/query subsets are defined in Section~5.2. The source-bank and metric details below use that protocol.

\subsection{Source adapter bank}
For backbone $b$, the bank contains all eligible fitted reference adapters
from the fixed source datasets:
\begin{equation}
\mathcal A_{\mathrm{src}}^{(b)}=
\bigcup_{D_i\in\mathcal D_{\mathrm{src}}}\mathcal A_i^{(b)},
\qquad N_{\mathrm{src}}^{(b)}=|\mathcal A_{\mathrm{src}}^{(b)}|.
\end{equation}
Here eligibility includes excluding source records from any subject present
in the target datasets. This exclusion is applied across related cohorts,
including ISRUC-S1/S3, before reference-adapter fitting. No target dataset
contributes adapters to the bank. There is no fixed bank-size cap or per-dataset
quota. After eligibility filtering, the bank contains 311 reference
adapters per backbone. Training uses dataset-balanced sampling so that large source datasets
do not dominate the adapter batches.

\subsection{Baseline interfaces}
The EEG-only baseline evaluates $h(f(X^{\mathrm{EEG}}))$. Mean-source transfer averages compatible fitted source vectors,
\begin{equation}
\widehat\omega_{\mathrm{mean}}=\frac1n\sum_{j=1}^n\omega_j^*.
\end{equation}
Nearest-source transfer selects a source adapter for reuse, whereas Direct MLP directly maps the condition to adapter weights. The controlled adapter comparisons are grouped by frozen backbone; native multimodal models remain separate cross-paradigm references.

\subsection{Reported metrics}
Let $N_{ab}$ count predictions with true class $a$ and predicted class $b$, with $n_a=\sum_bN_{ab}$, $\widehat n_b=\sum_aN_{ab}$, and $N=\sum_an_a$. For $K$ classes with positive support, balanced accuracy is
\begin{equation}
\text{B-ACC}=\frac1K\sum_{c=1}^K\frac{N_{cc}}{n_c}.
\end{equation}
Equal class weighting distinguishes balanced accuracy from ordinary accuracy, which weights the class recalls by their observed prevalences. The displayed result tables express B-ACC as a percentage; differences between those values are percentage points.

Weighted F1 uses the class-support weights:
\begin{equation}
\mathrm{WF1}=\sum_{c=1}^K\frac{n_c}{N}\mathrm{F1}_c,
\qquad \mathrm{F1}_c=\frac{2N_{cc}}{n_c+\widehat n_c},
\end{equation}
for nonzero denominators. Cohen's kappa adjusts observed agreement for agreement induced by the marginal label frequencies:
\begin{equation}
p_o=\frac{\sum_cN_{cc}}N,\quad
p_e=\sum_c\frac{n_c\widehat n_c}{N^2},\quad
\kappa=\frac{p_o-p_e}{1-p_e}.
\end{equation}
These metrics are not determined by B-ACC alone. Consequently, no kappa or weighted-F1 values are inferred from the B-ACC tables in this appendix. Likewise, the binary-task AUROC and AUPRC require score-level predictions; class-balanced accuracy cannot reconstruct either curve.

\subsection{Seed variation and pairwise comparisons}
Tables~3 and~5 report means and standard deviations over five seeds, matched within each respective comparison. The main-comparison and ablation batches are independent; seed matching does not extend across them. The plotted whiskers reproduce these reported SDs. A standard deviation describes seed-to-seed variation and is distinct from a standard error or confidence interval.

For two matched seed scores $x_s,y_s$, the difference is $d_s=x_s-y_s$. Its variance satisfies
\begin{equation}
\operatorname{Var}(d)=\operatorname{Var}(x)+\operatorname{Var}(y)-2\operatorname{Cov}(x,y).
\end{equation}
Marginal SDs do not determine the covariance. Therefore the derived mean gains in Appendix~\ref{sec:app_datasets} are shown without newly inferred paired error bars. The reported SDs of ablation decreases remain those given in Table~5.

\subsection{Aggregation}
The six-target comparisons average the 18 displayed backbone--dataset means with equal weight. This gives equal weight to each displayed setting, not to each participant or window. Within a target, the backbone groups reuse the target dataset and are not independent participant cohorts. The CBraMod ablation summary in Table~5 averages results over all nine target datasets, including CAP, DREAMER, and BCI-IV, yielding 73.41 for Full ZeroMAG. Its scope differs from the six-target comparison, and its experimental batch also differs from the nine-target main comparison (70.01\%); see Appendix~\ref{sec:app_batch_scope}.

Recovery ratios and score differences answer different questions. A percentage-point difference measures absolute movement in B-ACC; a recovery percentage divides that movement by a supervised-reference gain. The ratio-of-aggregate-gains convention in Appendix~\ref{sec:app_datasets} makes the 93.50\% summary explicit and separates it from the average of row-wise percentages.

\section{Generalization and Data Efficiency}
\label{sec:app_robustness}
\subsection{Modality holdout}
The generalization experiments summarized in the main text consider unseen combinations of known modalities and modality types in $\mathcal{M}$ that are excluded from generator training. ZeroMAG recovers 89.7\% and 82.6\% of the supervised multimodal gain, respectively. Fig.~5 compares the two recovery rates.

The type-holdout summary is 7.1 percentage points lower in recovery rate than the combination-holdout summary. This difference is in recovery percentage, not in balanced-accuracy percentage points. Both refer to the supported modality interface. They do not establish compatibility with an arbitrary signal type for which the model defines no encoder, identity embedding, or adapter block.

\subsection{Calibration data}
Performance saturates at approximately 64 unlabeled calibration windows in the experiment summarized in the main text. This concerns the amount of data used to form the condition, rather than a target optimization batch size: there is no target-side fitting step. Once constructed, the condition guides generation of the adapter used on the disjoint query subset.

A window count describes the number of calibration units rather than a universal duration in seconds. Window duration and signal sampling rates differ across physiological tasks. The reported saturation point therefore characterizes the evaluated calibration setting and does not imply that all recordings or tasks require exactly the same observation duration.

\subsection{Source-bank size}
Using 50\% of the source adapter bank retains 97.2\% of the full-bank multimodal gain in the reported data-efficiency experiment. This is a 2.8\% reduction in the reference gain, not a 2.8-point reduction in B-ACC. The denominator is the full-bank multimodal gain.

Together, the calibration and bank-size summaries concern two different resource requirements. Calibration windows are collected from a new target recording at deployment. Source adapters are fitted and organized offline. Reducing one does not directly reduce the other, and the reported summaries do not provide a joint optimum over calibration budget and bank size.

\FloatBarrier
\section{Comparison with Related Methods}
\label{sec:app_limitations}

\subsection{Extended related work}
\label{sec:app_related}

\paragraph{Multimodal physiological models}
SleepFM and PhysioOmni learn multimodal physiological representations through
pretraining~\citep{sleepfm2024,physioomni}, whereas Brant-X uses a pretrained
EEG foundation model to align EEG and other physiological signals and trains an
EXG encoder~\citep{brantx}.
ZeroMAG generates a plug-in adapter for a selected existing EFM, preserving its
frozen encoder and prediction head at target deployment.

\paragraph{Parameter-efficient and test-time adaptation}
Adapters, LoRA, and prompt tuning~\citep{adapters,lora,prompttuning} reduce the
number of trainable parameters but still optimize them on labeled target data.
Test-time and unsupervised domain adaptation~\citep{tent,shot} avoid
target labels but update parameters on target data, where hyperparameters and
stopping criteria are difficult to validate. ZeroMAG amortizes adaptation into
an offline-trained generator and performs no target-side gradient step.

\paragraph{Parameter generation}
Hypernetworks~\citep{hypernetworks}
generate weights from task representations, shared hypernetworks generate
adapters for multi-task fine-tuning~\citep{generatedadapters}, and neural
network diffusion models parameters with latent diffusion~\citep{pdiff}.
Recent work applies parameter generation to zero-shot adaptation of large
language models: Text-to-LoRA generates LoRA adapters from natural-language
task descriptions~\citep{charakorn2025texttolora}, and Drag-and-Drop LLMs maps
batches of unlabeled task prompts to LoRA weights~\citep{liang2025dnd}.
ZeroMAG shares the goal of replacing per-target optimization with generation,
but differs in three respects: its condition is computed from unlabeled
physiological recordings rather than text; the generated adapter must
accommodate varying modality configurations within a shared block schema; and
generation combines parameter-level targets with functional agreement through
the frozen EFM.

\subsection{Adaptation mechanisms}

Table~\ref{tab:method_comparison} distinguishes methods by the object they adapt and the information used for adaptation. It is a methodological comparison; the methods listed here are not additional experimental baselines.

\begin{table}[!ht]
\centering\small
\caption{Adaptation mechanisms in related work. Target-time optimization and conditional parameter generation are separate design choices.}\label{tab:method_comparison}
\begin{tabular}{>{\raggedright\arraybackslash}p{.29\linewidth}>{\raggedright\arraybackslash}p{.28\linewidth}>{\raggedright\arraybackslash}p{.34\linewidth}}
\toprule Method family & Adapted object & Adaptation mechanism\\\midrule
Adapters~\citep{adapters} & Inserted modules & Optimize a small set of task-specific parameters.\\
LoRA~\citep{lora} & Low-rank weight updates & Learn an update while freezing the original weights.\\
Prompt tuning~\citep{prompttuning} & Continuous input prompts & Optimize prompt embeddings for the task.\\
Tent~\citep{tent} & Normalization parameters & Minimize prediction entropy on test inputs.\\
SHOT~\citep{shot} & Target feature extractor & Adapt with a fixed source classifier and unlabeled target data.\\
Hypernetworks~\citep{hypernetworks,generatedadapters} & Generated network or adapter weights & Predict parameters from a learned conditioning representation.\\
Neural network diffusion~\citep{pdiff} & Network parameters & Generate parameters through a learned latent diffusion model.\\
ZeroMAG & Multimodal adapter & Generate weights from unlabeled calibration signals and task context; keep the EEG model fixed.\\
\bottomrule
\end{tabular}
\end{table}

The closest connection is to adapter generation and diffusion over network parameters. ZeroMAG uses a variable-length semantic block sequence to represent companion-signal adapters, and evaluates decoded weights through the frozen EEG prediction path. Its target condition comes from the recording and supported task. The generators for the three EEG backbones are trained separately.

The method depends on the supplied frozen EFM and compatible prediction head, the quality of source reference adapters, and the coverage of the calibration data. The reported results concern these evaluated interfaces. They do not establish transfer of one generator to an arbitrary new architecture or a task outside the supported vocabulary.

\end{document}